\documentclass[letterpaper]{article} % DO NOT CHANGE THIS
\usepackage{aaai2026}  % DO NOT CHANGE THIS
\usepackage{times}  % DO NOT CHANGE THIS
\usepackage{helvet}  % DO NOT CHANGE THIS
\usepackage{courier}  % DO NOT CHANGE THIS
\usepackage[hyphens]{url}  % DO NOT CHANGE THIS
\usepackage{graphicx} % DO NOT CHANGE THIS
\usepackage{natbib}  % DO NOT CHANGE THIS AND DO NOT ADD ANY OPTIONS TO IT
\usepackage{caption} % DO NOT CHANGE THIS AND DO NOT ADD ANY OPTIONS TO IT
\usepackage{algorithm}
\usepackage{algorithmic}

\usepackage{amsmath}
\usepackage{booktabs} 
\usepackage{color} % -- This package is specifically forbidden - remove in final version
\usepackage{colortbl}
\usepackage{enumitem}
\usepackage{makecell}
\usepackage{multirow}
\usepackage{xcolor}
\usepackage{tabularx}
\usepackage{tcolorbox}

\definecolor{50}{HTML}{c6dbef}
\definecolor{25}{HTML}{deebf7}
\definecolor{10}{HTML}{eff6fc}
\definecolor{0}{HTML}{f7f7f7}

\definecolor{502}{HTML}{2171b5}
\definecolor{252}{HTML}{4292c6}
\definecolor{102}{HTML}{6baed6}
\definecolor{02}{HTML}{636363}

\usepackage{newfloat}
\usepackage{listings}
\DeclareCaptionStyle{ruled}{labelfont=normalfont,labelsep=colon,strut=off} % DO NOT CHANGE THIS
\floatstyle{ruled}
\newfloat{listing}{tb}{lst}{}
\floatname{listing}{Listing}
\nocopyright
\title{Participatory Moral AI Is Not Neutral:\\ The Invisible Hand of Developers}

\author{
    Taenyun	Kim\textsuperscript{\rm 1},
    Edyta Bogucka\textsuperscript{\rm 2},
    Daniele Quercia\textsuperscript{\rm 2,3}\\
}
\affiliations{
    \textsuperscript{\rm 1}Michigan State University, US\\
    \textsuperscript{\rm 2}Nokia Bell Labs, Cambridge, UK\\
    \textsuperscript{\rm 3}Politecnico di Torino, Italy\\
    kimtaeny@msu.edu, edyta.bogucka@nokia-bell-labs.com, daniele.quercia@nokia-bell-labs.com
}

\usepackage{bibentry}

\begin{document}

\maketitle

\begin{abstract}
    As AI systems make more morally loaded decisions across society, one response has been moral preference elicitation. In this approach, researchers poll participants on hypothetical dilemmas and use the aggregated votes to train a policy that an AI model then applies at scale. Before any vote is cast, developers make three key choices in the moral AI elicitation pipeline: feature scoping, voter sampling, and question framing. In other words, they decide which features go to a vote, which voters to include, and how to present the question. These choices are often opaque, undocumented, and treated as technical details rather than normative ones. We examine each of these choices within a common empirical study and show that each can shape the preferences produced by moral AI elicitation. Across two phases (N = 809) in three deployment contexts (i.e., AI kidney allocation, AI agents simulating absent workers, and generative AI depictions of the deceased),  we examine the three main stages of the moral AI elicitation pipeline. First, morally relevant features shift across contexts. This suggests that feature schemas should not be assumed to transfer across deployment domains. Second, preferences differ by political ideology for roughly one-third of features, with some differences reversing direction. The ideological composition of the voter pool can therefore affect the resulting aggregated preference profile. Third, the wording of the elicitation question can narrow or widen ideological gaps by up to a full scale point. The framing conditions also change how moral foundations are associated with participants' judgments. Taken together, these findings suggest that voting-based alignment cannot deliver fair or transparent AI by aggregation alone; at minimum, each stage of the moral AI elicitation pipeline should be audited and disclosed.
\end{abstract}

% Uncomment the following to link to your code, datasets, an extended version or similar. You must keep this block between (not within) the abstract and the main body of the paper.

\begin{links}
    \link{Project website}{https://social-dynamics.net/moral-ai}
    %\link{Code}{https://aaai.org/example/code}
    %\link{Datasets}{https://aaai.org/example/datasets}
    %\link{Extended version}{https://aaai.org/example/extended-version}
\end{links}

%%%%%% SECTIONS %%%%%% 
\section{Introduction}
\label{sec:introduction}

AI systems are increasingly used to make high-stakes decisions in areas such as healthcare, employment, and criminal justice \cite{chan2024should,kim2025we}. Aligning these systems with human moral values in a way that is fair, accountable, and transparent has therefore become a central problem~\cite{boerstler2024stability, freedman2020adapting, Kneer2025hard}. To address it, \textit{moral preference elicitation} now stands as a common approach: researchers present participants with hypothetical moral dilemmas, ask them a question about what the AI system should do, and use these responses to train models that produce a decision policy~\cite{boerstler2024stability, freedman2020adapting, Kneer2025hard, noothigattu2018voting, awad2018moral}. The approach aims to democratize alignment by grounding model design in the preferences of a broad and representative public~\cite{simson2025preventing, dahl2020facilitating}.

In practice, however, before any question is asked, developers make three key choices that shape what the final policy can capture: \textit{feature scoping} (which features are included), \textit{voter sampling} (which participants are included), and \textit{question framing} (how the question is asked). These choices are rarely documented, and are typically treated as technical rather than moral decisions. Each of these choices can shape what is ultimately presented as an ``aggregated moral preference''. If the selected features, participant population, or question wording affect the results, the resulting policy reflects upstream developer decisions as well as participants' expressed values. We examine these three choices within a common empirical study, using one research question for each stage:

\begin{itemize}[leftmargin=2.85em, labelsep=0.5em, label={}]
    \item[\textbf{RQ\textsubscript{1}:}] \textit{Feature scoping.} Do people consider the same features across different use cases?
    \item[\textbf{RQ\textsubscript{2}:}] \textit{Voter sampling.} Do moral preferences vary with political ideology?
    \item[\textbf{RQ\textsubscript{3}:}] \textit{Question framing.} Does framing change moral preferences?
\end{itemize}

To address these questions, we conducted a study with participants holding different political views across three AI use cases: AI kidney allocation (KIDNEY), AI agents simulating absent workers (WORK), and generative AI content of the deceased (GEN). These cases vary in the severity of harm and everyday frequency of encounter, from rare high-stakes allocation to common workplace decisions and emerging questions about how AI should represent the deceased. We ran two phases for each use case. In Phase 1 ($N = 150, 149, 150$ per use case), participants identified which features should or should not matter. In Phase 2 ($N = 120$ per use case), participants judged whether each feature should count in favor, count against, or not count at all, under one of three conditions: a control, World\mbox{-}You\mbox{-}Want (a structural perspective~\cite{jaques2025moral}), or Could-Be-You (a perspective-taking approach based on Rawlsian theory~\cite{rawls1971theory, huang2019veil, bruno2024moral}). We then classified features as relevant, irrelevant, or divisive, and tested whether political ideology and framing changed these judgments. This analysis produced three main findings:

\begin{itemize}[leftmargin=1.3em, labelsep=0.5em, label={}]
    \item[1)]\textbf{Moral feature relevance is context-specific (\S\ref{dis:gen:diff}).} Relevant features vary by use case. KIDNEY emphasizes distributive justice and medical utility. WORK focuses on accountability and legitimacy. GEN centers on consent and dignity. The relevant feature sets differed across the three contexts. Choosing which features to include is therefore a moral decision that defines the boundaries of the elicitation process.

    \item[2)]\textbf{Voter sampling can change the aggregated preference profile (\S\ref{dis:gen:political}).}  Political ideology is associated with different evaluations for roughly one-third of features, including some differences that reverse direction. In KIDNEY, conservatives place greater weight on proportionality and utility, while progressives place greater weight on equality. In WORK, conservatives emphasize employer loyalty and contractual duty. In GEN, conservatives place greater weight on the dignity of the deceased, while progressives are more permissive toward creative use. The ideological composition of the voter pool can therefore affect the preferences represented in the aggregate.

    \textbf{3) Question framing changes ideological gaps and value pathways (\S\ref{dis:gen:framing}).} Framing can change results by up to one scale point, narrowing some ideological gaps while widening others. The framing conditions also change how moral foundations are associated with participants' judgments. Question framing should therefore be treated as a consequential design choice rather than a neutral elicitation tool.
\end{itemize}

Taken together, these findings show that voting-based alignment does not simply recover a pre-existing set of public moral preferences: its outputs are shaped by upstream choices about what is put to a vote, whose preferences are represented, and how those preferences are elicited. This paper makes three contributions. First, it conceptualizes feature scoping, voter sampling, and question framing as normative developer choices in moral AI elicitation. Second, it provides comparative empirical evidence about each choice across three deployment contexts. Third, it translates these findings into a sensitivity audit that makes upstream choices visible and contestable (\S\ref{dis:gen:design}), while examining the limits of aggregation even when those choices are disclosed (\S\ref{dis:gen:ethical}).
\section{Related Work}
Next, we review prior work on voting-based AI alignment and the three developer choices that structure it: \textit{feature scoping}, \textit{voter sampling}, and \textit{question framing}.

\subsection{Voting-Based AI Alignment and Its Assumptions}

Aligning AI systems with human moral values is a central concern for AI safety research~\cite{gabriel2020artificial, Kneer2025hard}. One common approach is \textit{moral preference elicitation}, in which researchers present hypothetical dilemmas, collect participants' responses, and aggregate them into a decision policy that a model applies at scale~\cite{noothigattu2018voting, awad2018moral, freedman2020adapting, keswani2025can}. This approach is often framed as democratizing alignment by replacing developers' private judgments with public input~\cite{simson2025preventing, dahl2020facilitating}.

Yet developers still decide what participants evaluate, who participates, and how questions are posed. We call these choices \textit{feature scoping}, \textit{voter sampling}, and \textit{question framing}. Claims that aggregation neutrally recovers public values assume that features transfer across contexts, voter composition has limited influence, and wording reveals rather than shapes values. Prior work challenges these assumptions: aggregation can marginalize minority views~\cite{feffer2023moral}, while demographic groups prioritize ethical values differently~\cite{jakesch2022different}. Anthropic's Constitutional AI illustrates the issue: although 1000 Americans contributed to a model constitution, developers still selected participants, statements, and training principles~\cite{ganguli2023collective}. More fundamentally, aggregated preferences cannot determine which views deserve authority, how conflicts should be resolved, or what protections should constrain majority decisions~\cite{Salloch2015, Harris_2020, Baum2025, ZhiXuan2024}. Taken together, this suggests preferences should be treated as one input into moral reasoning rather than as a substitute for it. Our study does not attempt to resolve this problem. Instead, it examines how upstream design choices further condition what is presented as an aggregated preference.

\subsection{Moral Preferences Across AI Use Cases}
Most moral preference elicitation studies focus on a small set of use cases that tend to share three properties: they involve severe harm if the AI fails or is misused, they are rare or one-off situations, and they concern decisions that humans have traditionally made case by case. Organ allocation and autonomous vehicle scenarios are the clearest examples of this pattern, with concerns centered on life-and-death tradeoffs and the allocation of scarce resources~\cite{noothigattu2018voting, awad2018moral}. More frequent, lower-stakes cases such as content moderation, ad targeting, and consumer AI have received far less attention, yet they raise different moral concerns around transparency, consent, and accountability~\cite{jakesch2022different}.

Within a use case, feature scoping can follow two approaches: top-down and bottom-up. In a top-down approach, developers select features before recruiting participants, drawing on existing literature or industry guidelines. In a bottom-up approach, participants themselves identify which considerations they find morally relevant~\cite{valueSensitiveAI2023}. Most studies have used the top-down approach. For example, \citet{awad2018moral} reviewed the Western academic literature to identify 18 features describing the people affected by an autonomous vehicle's decision, including their gender, age, and physical fitness. Critics argue that this forced-choice format narrows the ethical problem by preselecting which people, outcomes, and trade-offs participants can consider, while omitting factors such as uncertainty and how likely each outcome is~\cite{Dewitt2019, Etienne2021, DrFreitas2021, Schuessler2023}. Similarly, \citet{jakesch2022different} drew on industry ethics guidelines to identify 12 values relevant to responsible AI in ad targeting, including safety and fairness. In both cases, what goes to a vote was decided before any participant was recruited, and researchers have called for such decisions to be documented and opened to scrutiny~\cite{feffer2023moral,situatedValues2025}.

\subsection{Group Differences in Moral Preferences}

Differences in moral judgment across groups are well documented. In moral preference elicitation tasks, participants with different demographic, AI literacy, or political backgrounds often reach different conclusions \cite{graham2009liberals, atari2023morality}. This diversity poses a challenge for systems that attempt to learn a single set of ``universal'' moral preferences~\cite{gabriel2020artificial, jakesch2022different}. Disagreements arise at two levels: people differ in which features they consider relevant ~\cite{keswani2025can}, and they differ in how they weight those features~\cite{brugman2024effects}. For example, when building AI systems, practitioners flag safety and privacy as more relevant than the general public does~\cite{jakesch2022different}, and developers weight politeness over the straightforwardness that the users they build for actually want~\cite{ranjit2026automating}.

These differences are especially pronounced across political ideology. Moral Foundations Theory (MFT) suggests that progressives emphasize \textit{care} and \textit{equality}, whereas conservatives place greater weight on \textit{proportionality}, \textit{authority}, \textit{loyalty}, and \textit{purity}~\cite{graham2009liberals, atari2023morality}. These moral differences shape how people interact with AI systems~\cite{brailsford2024exploring}. In automated vehicle scenarios, for example, sacrificial preferences vary by political ideology~\cite{awad2018moral}. Taken together, these findings suggest that the ideological composition of a voter pool can affect the values represented in the aggregated preference profile.

\subsection{Moral Preference Influenced by Moral Framing}

Moral framing is the selection and emphasis of particular aspects of a morally charged situation, shaping how people interpret and evaluate it~\cite{brugman2024effects,semetko2000framing}. It can influence preferences in two ways: through question framing, which emphasizes particular moral concerns, and through response framing, which constrains the responses participants can express.

Question framing can shift preferences by changing how a decision is presented or which considerations participants are prompted to weigh~\cite{feinberg2019moral,gamson1989media,rehren2021moral}. When evaluating an isolated decision, people respond differently when identical outcomes are framed as lives saved rather than lives lost~\cite{McDonald_2021}. Socratic questions can also prompt participants to justify and reconsider that individual decision~\cite{torabizadeh2018impacts}. Other frames shift attention from the isolated decision to the consequences of adopting it as a general policy. \citet{jaques2025moral} argues that single-case judgments can invite individual bias and obscure the effects of scaling a choice, and instead proposes asking, ``What kind of world would I be creating if this became the AI policy?''~\cite{jaques2025moral}. The veil of ignorance similarly asks people to evaluate a policy without knowing their own position in the resulting society~\cite{rawls1971theory,huang2019veil,huang2021veil,bruno2024moral}. Such frames have been associated with less self-serving reasoning, greater attention to stakeholder perspectives, and reduced political disagreement~\cite{huang2019veil,huang2021veil,bruno2024moral,franks2019economic,whitmarsh2017tools,bloemraad2016rights}.

Response framing instead limits which preferences participants can express. When forced to choose which demographic group to spare, participants appeared to support unequal treatment; when given an equal-treatment option, most selected it~\cite{Bigman2020}.

\smallskip
\noindent\textbf{Research Contribution.}  Prior research shows that feature scoping, voter sampling, and question framing can each shape moral judgments. Building on this work, we examine the three choices as successive stages of moral AI preference elicitation within a common empirical project. Our contribution is an integrated, cross-context account of how developer choices enter the elicitation process: we identify context-specific feature scopes, estimate ideological differences in feature evaluations, and test the effects of two explicit framing interventions. We conduct these analyses across three deployment contexts that differ in moral stakes and frequency of encounter. We then use the combined evidence to develop a sensitivity audit and recommendations for voting-based alignment (\S\ref{dis:gen:design}).
\begin{figure*}
    \centering
    \includegraphics[width=\textwidth]{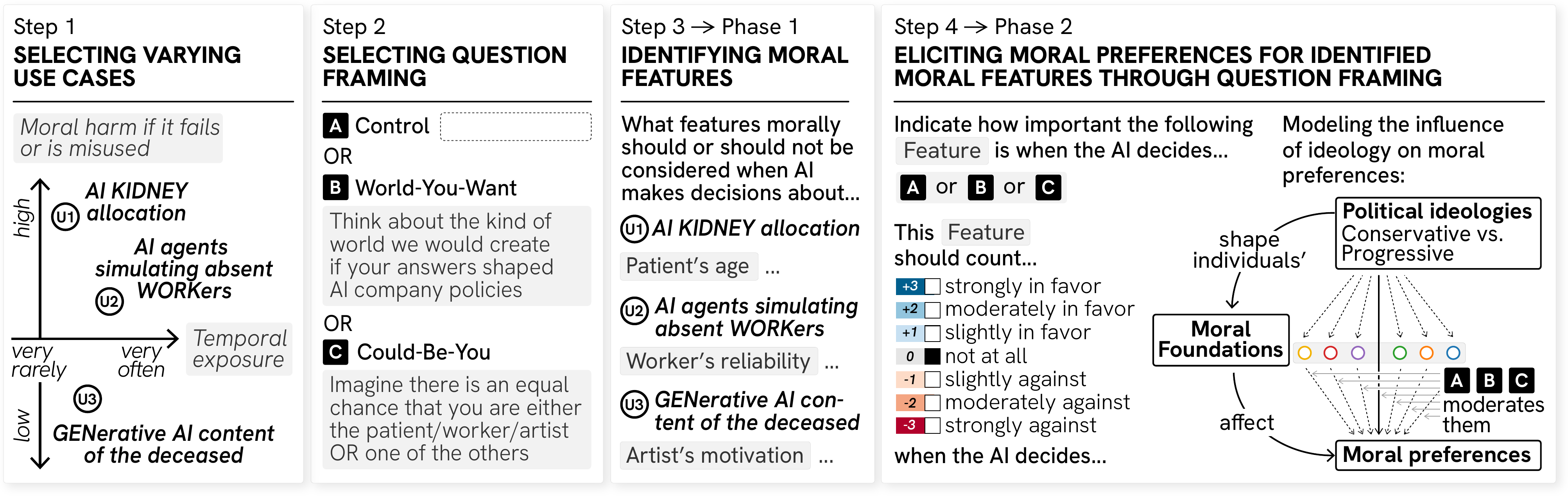}
        \caption{\textbf{Overview of our four-step methodology.}
        \emph{Step 1.} We selected three use cases---AI kidney allocation (KIDNEY), AI-simulated workers (WORK), and generative AI content of the deceased (GEN)---based on harm impact and likelihood of encounter, capturing rare high-stakes to common but less harmful cases.
        \emph{Step 2.} We used three framing conditions---(A) \emph{Control} (baseline), (B) \emph{World-You-Want} (societal consequences), and (C) \emph{Could-Be-You} (perspective-taking)---capturing structural and perspective-taking prompts to examine how different moral lenses shape moral preference elicitation.
        \emph{Step 3.} In Phase 1, participants identified moral features relevant to AI decision-making for each use case.
        \emph{Step 4.} In Phase 2, we elicited moral preferences for these features from a new sample using a 7-point scale ($-3$ to $+3$) across the three framings in Step 2. We then modeled how political ideology influences these preferences through moral foundations, with framing condition as a moderator.}
    \label{fig:teaser}
\end{figure*}

\section{Methods}

We study moral preferences and the effects of question framing across three use cases (\S\ref{s:use-case-selection}), three framing conditions (\S\ref{s:moral-framing}), and two phases. Phase 1 identifies candidate moral features (\S\ref{section:study1}). Phase 2 evaluates the moral importance of these features under different framing conditions (\S\ref{section:study2}).

\subsection{Use Case Selection}
\label{s:use-case-selection}

We select three use cases based on two criteria (Figure \ref{fig:teaser}, Step 1): impact (the scale of harm if the system fails or is misused) and temporal exposure (how often people face the dilemma). Together, they cover a range of ethical settings, from high-stakes allocation to everyday AI use:

\begin{itemize}[leftmargin=1.3em, labelsep=0.5em, label={}]
    \item[1)]\textbf{AI kidney allocation (KIDNEY)} is a life or death dilemma rooted in distributive justice. It raises questions about fairness and the role of AI in allocating scarce resources~\cite{keswani2025can}. While common in moral elicitation research~\cite{keswani2025can,boerstler2024stability,freedman2020adapting}, such cases are rare in daily life~\cite{jaques2025moral,nguyen2022mapping}. 
    
    \item[2)]\textbf{AI agents simulating absent workers (WORK)} captures a more common setting~\cite{yudkin2025large}. It looks at remote workers who use AI to simulate activity. Concerns include deception, work norms, and AI compliance with questionable instructions~\cite{boland2025moral}.
    
    %The main concerns are deception, work norms, and whether AI should follow questionable instructions~\cite{boland2025moral}.
     
    \item[3)]\textbf{Generative AI content of the deceased (GEN)} raises issues of consent after death, emotional harm, and commercial use~\cite{buben2025replacement, danaher2025mvpp}. These systems may change how people remember the dead or weaken a person’s unique identity~\cite{danaher2024scarcity, lazaridis2025doppelgangers}.
\end{itemize}

\subsection{Question Framing Selection}
\label{s:moral-framing}

We selected three question framing conditions for the elicitation task in Phase 2:

\begin{itemize}[leftmargin=1.9em, labelsep=0.5em, label={}]
    \item[(A)]\textbf{Control} uses no additional question framing prompt (Figure \ref{fig:teaser}, Step 2A).
    
    \item[(B)]\textbf{World-You-Want} draws on Jaques's structural perspective~\cite{jaques2025moral}. It frames the elicitation question by prompting participants to consider long-term societal effects and AI policy. Participants read: ``Think about the kind of world we would create if your answers shaped AI company policies'' (Figure \ref{fig:teaser}, Step 2B).
    
    \item[(C)]\textbf{Could-Be-You} draws on Rawls's veil of ignorance~\cite{rawls1971theory, huang2019veil, huang2021veil}. It frames the elicitation question by prompting participants to take the perspective of any affected stakeholder~\cite{bruno2024moral}. For example: ``Imagine there is an equal chance that you are either one of the patients'' (Figure \ref{fig:teaser}, Step 2C).
\end{itemize}

\subsection{Phase 1: Identifying Moral Features}
\label{section:study1}

\smallskip\noindent\textbf{Participants.}
We recruited participants via Prolific at a rate of at least 8 USD/hour, following sample sizes established in prior work using the same two-phase design~\cite{keswani2025can, chan2022features} ($N_{KIDNEY}=150$, $N_{WORK}=149$, $N_{GEN}=150$). We used quota sampling to balance political ideology (conservative, moderate, progressive) within each use case. Full demographic breakdowns, including age, gender, race/ethnicity, and education level, are reported in Table \ref{tab:demographics_s1} in Supplementary Materials \ref{app:demographics}.

\smallskip\noindent\textbf{Procedure.}
After consent and a use case introduction, participants adopted a ``moral point of view''. They listed 5 features that morally \textit{should} be considered and 5 that \textit{should not}. They provided a justification for each feature. For features that should be considered, they also gave example levels (e.g., ``young'' and ``old'' for ``age'') (Figure \ref{fig:teaser}, Step 3; Figure \ref{app:instructions} in Supplementary Materials \ref{app:phase_instructions}).

\smallskip\noindent\textbf{Analysis.} To identify moral features, we used LLM-assisted coding with \texttt{GPT-4o-mini}. We selected this model for its low cost and high performance on instruction-following tasks~\cite{rytting2023codingsocialsciencedatasets, ranjit2024oathframescharacterizingonlineattitudes}. We ran it once per participant response at temperature 0 using the prompt in Supplementary Materials~\ref{app:llm-prompt}. The LLM extracted feature names from responses, summarized participants' justifications, and assigned each feature one of four relevance labels (``should'', ``should not'', ``mixed'', ``unclear'').

To assess LLM output quality, two authors independently coded a random sample of 100 responses, reviewing both feature names and relevance labels. Agreement between the authors and the LLM was 85\% on feature names and 97\% on relevance labels. To refine feature names across all outputs, authors conducted five calibration rounds following standard qualitative methods~\cite{charmaz2015grounded, oktay2012grounded}. They resolved disagreements by consensus (with a third author as tie-breaker) and recorded all decisions. This included, for example, merging synonymous codes (e.g., ``geographic location'' and ``travel time'' into ``geographic proximity'').

\bigskip
\noindent\textbf{Feature inclusion threshold as a developer choice.} We computed feature prevalence using binary indicators (1 if mentioned by a participant). Following prior work~\cite{keswani2025can}, we retained the top 30 to 35 most prevalent features, excluding those mentioned by fewer than 4 participants ($\approx$2.5\% of our sample). For example, in the KIDNEY scenario, \textit{taxpayer status} and \textit{favorite sports team} were each mentioned by only 1 participant and thus excluded. We treat this threshold as a design choice within \textit{feature scoping} that may exclude minority perspectives.

\subsection{Phase 2: Eliciting Moral Preferences Under Framing}
\label{section:study2}

\smallskip\noindent\textbf{Participants.}
We recruited 360 US participants (120 per use case) via Prolific at a rate of at least 8 USD/hour. We randomly assigned participants to the Control, World-You-Want, or Could-Be-You condition, with equal representation of conservatives and progressives per condition. Table \ref{tab:demographics_s2} in Supplementary Materials \ref{app:demographics} reports full demographic breakdowns, including age, gender, race/ethnicity, education, AI literacy, religiosity, and area of residence.

\smallskip\noindent\textbf{Procedure.}
After consent and a use case introduction, participants evaluated features from Phase 1 (Figure \ref{fig:teaser}, Step 4). Each feature appeared as a contrast (e.g., ``younger vs. older patients'' for KIDNEY, ``grief support vs. commercial use'' for GEN). Participants rated moral importance on a 7-point scale from $-3$ (\textit{count strongly against}) to $+3$ (\textit{count strongly in favor}), with $0$ indicating \textit{not count at all} (see Figures \ref{app:study2_kidney_treatment_appendix}, \ref{app:study2_work} and \ref{app:study2_art} in Supplementary Materials \ref{app:phase_instructions}). Unlike pairwise methods, which scale poorly with many features~\cite{boerstler2024stability, keswani2025can}, this scale allowed us to evaluate more than 30 features. This design supports analysis of how political ideology, moral foundations, and framing interact.

Participants also completed two validated measures. \textit{Moral Foundations} used the 36-item MFQ-2~\cite{atari2023morality}, which measures six foundations: Care, Equality, Proportionality, Loyalty, Authority, and Purity~\cite{graham2009liberals}. Responses used a 5-point scale (1: \textit{not at all}, 5: \textit{extremely well}). \textit{AI Literacy} used a 17-item scale~\cite{tully2025lower} with two components: technical understanding (10 items) and awareness of limitations and ethics (7 items). Each item was scored as correct or incorrect and scaled to $[0,1]$. We use AI literacy as a proxy for prior experience with AI. 

\smallskip\noindent\textbf{Analysis.}
We classified features into six categories based on relevance ($M_r, p_r$ via t-test vs.\ 0.5) and direction ($M_d, p_d$ via t-test vs.\ 0). We coded relevance as 0 (``not count at all'') or 1 (any other value). We used the raw scale for direction. Categories were:

\begin{itemize}[leftmargin=1.3em, labelsep=0.5em, label={}]
    \item[1)]\emph{Morally relevant and counting for}, if $p_r < 0.05$ and $M_r > 0.5$, and $p_d < 0.05$ and $M_d > 0$. \vspace{3pt}
    \item[2)]\emph{Morally relevant and counting against}, if $p_r < 0.05$ and $M_r > 0.5$, and $p_d < 0.05$ and $M_d < 0$. \vspace{3pt}
    \item[3)]\emph{Morally irrelevant}, if $p_r < 0.05$ and $M_r < 0.5$. \vspace{3pt}
    \item[4)]\emph{Morally relevant but directionally divisive}, if $p_r < 0.05$ and $M_r > 0.5$, and $p_d \geq 0.05$. \vspace{3pt}
    \item[5)]\emph{Morally divisive}, if $p_r \geq 0.05$. \vspace{3pt}
    \item[6)]\emph{Morally divisive with directional disagreement}, if $p_r \geq 0.05$ and $p_d \geq 0.05$.
\end{itemize}

We then ran moderated mediation analyses using Hayes's PROCESS Model 15 with 5,000 bootstrap samples (v4.3 for R)~\cite{hayes2017introduction}. These models test how political ideology affects moral foundations, how these foundations affect moral importance, and how framing moderates these relationships (see Figure~\ref{fig:study2_model} in Supplementary Materials \ref{app:model}). We control for sex (1 = female), age, education (1 = pre-college; 2 = college; 3 = advanced), ethnicity (1 = non-Hispanic White), religiosity (1 = not important; 5 = extremely important), area of residence (1 = rural), and AI literacy.
\section{Results}

\begin{figure}
    \centering
    \includegraphics[width=\columnwidth]{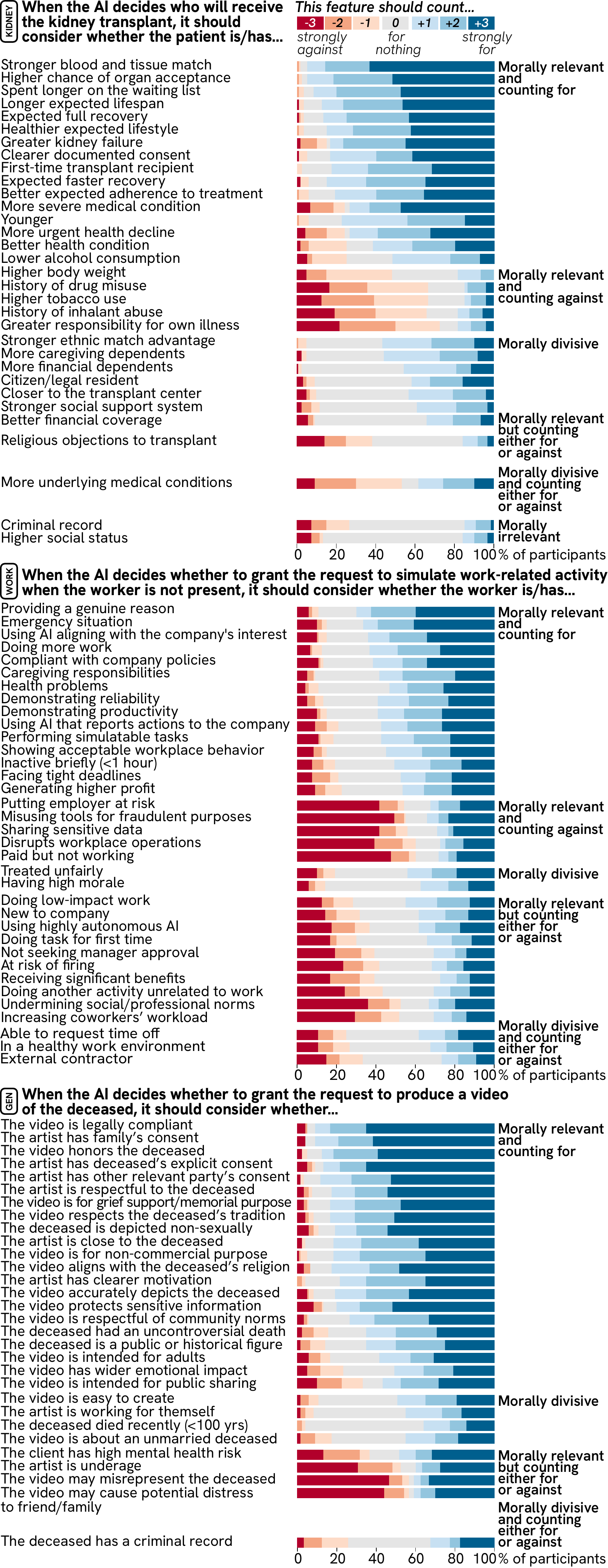}
    %\caption{\textbf{Distribution of ratings in Phase 2 for the AI kidney allocation (KIDNEY), AI-simulated workers (WORK), and Generative AI content of the deceased (GEN) use cases.} Results were aggregated across conditions, and features were ordered by moral feature category and by mean rating from highest to lowest.}
    \caption{\textbf{Phase 2 rating distributions for KIDNEY, WORK, and GEN use cases.} Features were ordered by moral feature category and mean rating, highest to lowest.}
    \label{fig:study2_cases}
\end{figure}

\subsection{RQ\textsubscript{1}. Do People Consider the Same Features Across Different Use Cases?}

The features participants considered morally relevant were
highly context-specific across KIDNEY, WORK, and GEN (Figure \ref{fig:study2_cases}).

In KIDNEY, participants focused on medical utility, such as patients' \emph{health status} ($47\%$) and \emph{chance of survival} ($35\%$). In WORK, they emphasized worker motivation (e.g., \emph{reason for request}, $32\%$) and organizational factors (e.g., company policy, $27\%$). In GEN, they highlighted ethical safeguards, including the \emph{deceased's consent} ($28\%$) and \emph{intended purpose} ($40\%$). Full results are reported in Table  \ref{tab:kidney-study1-should} in Supplementary Materials \ref{app:results_kidney}, Table \ref{tab:work-study1-should} in Supplementary Materials \ref{app:results_work}, and Table \ref{tab:art-study1-should} in Supplementary Materials \ref{app:results_art}.

Participants also showed strong agreement on features that should \emph{not} be considered. These were mostly demographic or socioeconomic attributes, including \emph{ethnicity} (KIDNEY: $66\%$; WORK: $18\%$; GEN: $12\%$), \emph{gender} (KIDNEY: $64\%$; WORK: $14\%$; GEN: $12\%$), and \emph{social status} (KIDNEY: $27\%$; WORK: $12\%$; GEN: $12\%$). Age was also rejected in WORK ($18\%$) and GEN ($13\%$), but $82\%$ of participants in KIDNEY identified it as morally relevant. Table \ref{tab:kidney-study1-shouldnt} in Supplementary Materials \ref{app:results_kidney}, Table \ref{tab:work-study1-shouldnt} in Supplementary Materials \ref{app:results_work}, and Table \ref{tab:art-study1-shouldnt} in Supplementary Materials \ref{app:results_art} provide further details.

Phase 2 shows that most features reached consensus on moral relevance (KIDNEY: $65\%$; WORK: $57\%$; GEN: $70\%$). These features were typically tied to direct utility. For example, in KIDNEY, a patient with \emph{a stronger blood and tissue match with a donor} was judged in favor ($M = 2.42$, $SD = 0.95$, $p_r < .001$, $p_d = .001$), whereas \emph{greater responsibility for one's illness} was judged against ($M = -1.10$, $SD = 1.69$, $p_r < .001$, $p_d < .001$).

Similar patterns appear in WORK and GEN. In WORK, a worker's \emph{emergency situation} was judged in favor ($M = 1.29$, $SD = 1.97$, $p_r < .001$, $p_d < .001$), while \emph{disrupted workplace operations} was judged against ($M = -0.84$, $SD = 2.30$, $p_r < .001$, $p_d = .003$). In GEN, requests for \emph{grief support or memorial purposes} ($M = 1.88$, $SD = 1.49$, $p_r < .001$, $p_d < .001$) or those that \emph{honored the deceased} ($M = 2.18$, $SD = 1.33$, $p_r < .001$, $p_d < .001$) were judged in favor.

Despite this agreement, many features remained divisive (KIDNEY: $31\%$; WORK: $43\%$; GEN: $30\%$). These features often had unclear or mixed implications. Participants agreed they were relevant but disagreed on direction, for or against. Examples include \emph{more underlying medical conditions} in KIDNEY ($M = -0.18$, $SD = 1.88$, $p_r < .001$, $p_d = 1.0$), \emph{increasing coworkers' workload} in WORK ($M = -0.57$, $SD = 2.15$, $p_r < .001$, $p_d = 1.0$), and \emph{risk of misrepresentation} in GEN ($M = -0.45$, $SD = 2.75$, $p_r < .001$, $p_d = 0.36$). Table \ref{tab:kidney_study2} in Supplementary Materials \ref{app:results_kidney}, Table \ref{tab:work-study2} in Supplementary Materials \ref{app:results_work}, and Table \ref{tab:art-study2} in Supplementary Materials \ref{app:results_art} provide further details.

\subsection{RQ\textsubscript{2}. Do Moral Preferences Vary With Political Ideology?}

Preferences differed by political ideology for roughly one-third of features, with some differences reversing direction. Consequently, changing the ideological composition of an aggregated sample could change the preference profile.

In the control condition, ideological differences appeared across all use cases. In KIDNEY, conservatives favored patients with \emph{expected full recovery} ($b = 1.08$, $95\%\,CI\,[0.12, 2.03]$) and penalized those with \emph{more underlying medical conditions} ($b = -1.71$, $95\%\,CI\,[-3.20, -0.23]$). In WORK, conservatives opposed requests from workers who were \emph{being paid but not working} ($b = -2.79$, $95\%\,CI\,[-5.58, -0.00]$). In GEN, they were less likely to approve requests when the video risked \emph{misrepresenting the deceased} ($b = -2.91$, $95\%\,CI\,[-5.62, -0.20]$). Table \ref{tab:mm_kidney} in Supplementary Materials \ref{app:results_kidney}, Table \ref{tab:mm_work} in Supplementary Materials \ref{app:results_work}, and Table \ref{tab:mm_art} in Supplementary Materials \ref{app:results_art} provide further details.

\subsection{RQ\textsubscript{3}. Does Question Framing Change Moral Preferences?}

\begin{figure*}[t]
    \centering
    \includegraphics[width=\textwidth]{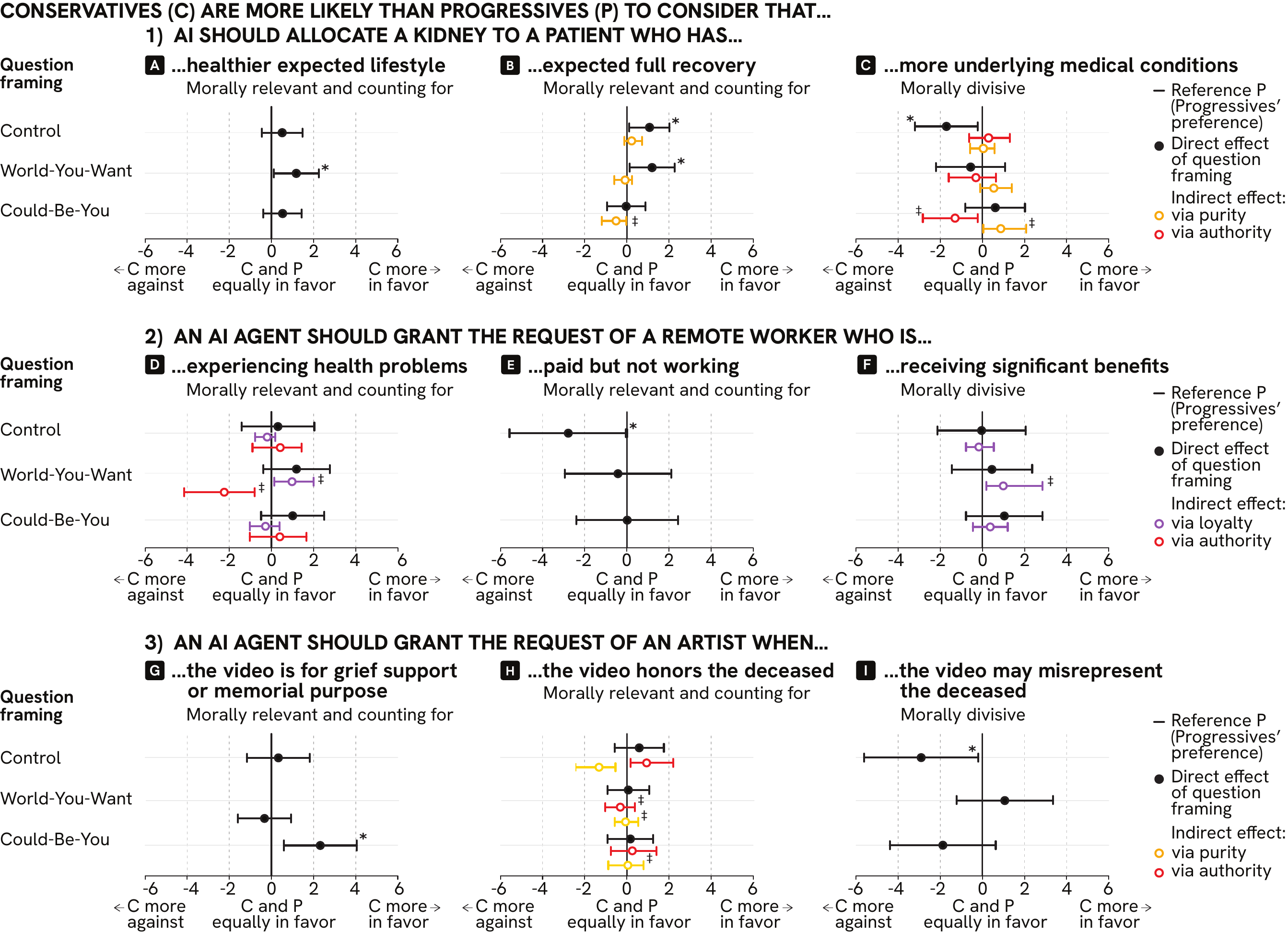}
    \caption{\textbf{Moderated mediation results for our three use cases: KIDNEY, WORK, GEN (Phase 2).} \textit{Note:} *The conditional direct effect for conservatives differed significantly from that for progressives. \textsuperscript{$\ddagger$}The index of moderated mediation was statistically significant (bootstrapped 95\% confidence interval excluding zero). Additional results are shown in Figure \ref{fig:study2:mm:kidney:appen} in Supplementary Materials \ref{app:results_kidney} and Figure \ref{fig:study2:mm:art:appen} in Supplementary Materials \ref{app:results_art}, available on the project website.} 
    \label{fig:results}
\end{figure*}

The third developer choice (\textit{question framing}) systematically shifts preferences. It can reduce some ideological gaps while increasing others. The magnitude of these shifts reaches up to one full scale point (Figure \ref{fig:results}).

\smallskip\noindent\textbf{Both Question Framings.} In KIDNEY, disagreement about patients with \emph{more underlying medical conditions} decreased under both framings (World-You-Want: $b = -0.56$, $95\%\,CI\,[-2.21, 1.08]$; Could-Be-You: $b = 0.61$, $95\%\,CI\,[-0.81, 2.02]$; Figure \ref{fig:results}C).

In WORK, conservatives' opposition to workers \emph{being paid but not working} was also reduced (World-You-Want: $b = -0.43$, $95\%\,CI\,[-2.96, 2.10]$; Could-Be-You: $b = 0.01$, $95\%\,CI\,[-2.41, 2.43]$; Figure \ref{fig:results}E).

In GEN, both framings reduced disagreement about videos that risk \emph{misrepresenting the deceased} (World-You-Want: $b = 1.06$, $95\%\,CI\,[-1.23, 3.35]$; Could-Be-You: $b = -1.88$, $95\%\,CI\,[-4.40, 0.63]$; Figure \ref{fig:results}I).

\smallskip\noindent\textbf{Could-Be-You Question Framing.} In KIDNEY, conservatives' preference for \emph{expected full recovery} (present in Control and World-You-Want) disappeared under Could-Be-You ($p = .95$). This change was associated with reduced influence of purity ($b= -0.51$, $95\%\,CI\,[-1.19, -0.03]$; $IMM=-0.74$, $95\%\,CI\,[-1.68, -0.07]$; Figure \ref{fig:results}B). The same condition also introduced tension between moral foundations. Authority predicted opposition to \emph{patients with more underlying conditions} ($b = -1.30$, $95\%\,CI\,[-2.83, -0.23]$), while purity predicted support ($b~= 0.87$, $95\%\,CI\,[0.04, 2.08]$; Figure \ref{fig:results}C). No comparable effects appeared in WORK. In GEN, however, Could-Be-You increased conservatives' support for requests related to \emph{grief support and memorial purposes} ($b = 2.32$, $95\%\,CI\,[0.94, 3.71]$, $p < .02$; Figure \ref{fig:results}G).

Response-time differences suggest that the framing changed how participants engaged with the task, although response time alone cannot identify the underlying cognitive process. In KIDNEY, participants in Could-Be-You took longer to respond ($EMM = 12.86$, $95\%\,CI\,[11.20, 14.77]$) than those in World-You-Want ($EMM = 9.10$, $p < .001$) or Control ($EMM = 9.43$, $p = .008$).
\footnote{The ANOVA used log-transformed values. Reported EMMs are back-transformed.}

\bigskip\noindent\textbf{World-You-Want Question Framing.} In KIDNEY, conservatives favored \emph{healthier lifestyles} more than progressives under World-You-Want ($b = 1.18$, $95\%\,CI\,[0.12, 2.25]$), but not in Control ($b = 0.51$, $95\%\,CI\,[-0.45, 1.47]$) or Could-Be-You ($b = 0.53$, $95\%\,CI\,[-0.39, 1.44]$; Figure \ref{fig:results}A).

In WORK, World-You-Want increased the role of loyalty. Conservatives with stronger loyalty values showed greater support for workers \emph{experiencing health problems} or \emph{receiving significant benefits} ($b = 0.98\text{--}0.99$, $IMM = 1.16\text{--}1.18$; Figure \ref{fig:results}D, F). At the same time, authority predicted opposition, creating tension between foundations.

In GEN, World-You-Want reduced this type of tension. In the Control condition, authority supported requests that \emph{honor the deceased}, while purity opposed them. Under World-You-Want, neither foundation had a significant effect (Figure \ref{fig:results}H).
\bigskip
\section{Discussion}

We discuss our three main findings (\S\ref{dis:gen:diff}--\S\ref{dis:gen:framing}), translate them into recommendations for voting-based alignment (\S\ref{dis:gen:design}), and reflect on the limits of aggregation as an alignment method (\S\ref{dis:gen:ethical}).

\subsection{Finding 1: Moral Feature Relevance Is Context-Specific}
\label{dis:gen:diff}

The first developer choice in the elicitation pipeline (which features are put up for a vote) is often treated as a pre-processing step, but our results show it functions as a per-deployment moral decision: the relevant feature sets differed across the three contexts.

In KIDNEY, participants focused on medical utility, especially the chance of survival, which reflects the distributive justice structure of the task~\cite{keswani2025can}. In WORK, they emphasized worker motivation and organizational factors, which align with concerns about deception, accountability~\cite{yudkin2025large}, and the legitimacy of commands~\cite{boland2025moral}. In GEN, attention shifted to the purpose of the video and the deceased's consent, which mirrors ongoing debates about posthumous representation~\cite{buben2025replacement, danaher2025mvpp}.

Features became divisive when their implications were unclear. In KIDNEY, participants disagreed on whether underlying conditions should prioritize patients based on need or penalize them due to poor prognosis. This reflects a tension between utility and equity~\cite{keswani2025can}. In WORK, some participants justified AI actions when harm seemed minimal~\cite{blanken2015meta}, while others rejected such actions as inherently wrong. In GEN, disagreement centered on the risk of misrepresentation, reflecting a conflict between dignity and harm prevention~\cite{buben2025replacement, danaher2025mvpp}.

Across all use cases, participants generally rejected socio-demographic features as morally relevant unless they were necessary for the decision (e.g., age in KIDNEY). This suggests that participants view such features as discriminatory unless justified~\cite{keswani2025can}. Because feature sets differed across contexts, developers delimit what the elicitation process can express before votes are collected.

\subsection{Finding 2: Voter Pool Composition Can Change Aggregated Preferences}
\label{dis:gen:political}

Preferences differed by political ideology for roughly one-third of features, with some differences reversing direction. Consequently, changing the ideological composition of an aggregated sample could change the preference profile.

These differences reflect distinct value priorities. In KIDNEY, conservatives favored allocating resources to patients with higher chances of recovery and opposed allocating them to patients with more underlying conditions. This pattern aligns with proportionality-based reasoning, while progressive responses reflect stronger emphasis on equality~\cite{graham2009liberals}. In WORK, conservatives more strongly opposed requests to simulate work activity when employees were being paid but not working. This suggests greater emphasis on loyalty and contractual obligation~\cite{graham2009liberals, koleva2012tracing}. In GEN, conservatives more strongly opposed requests that risk misrepresenting the deceased, while progressives were more permissive. This pattern may reflect broader political differences in expressive norms~\cite{kfrerer2021politics}, and ongoing debates about digital remains~\cite{danaher2024scarcity, lazaridis2025doppelgangers, buben2025replacement}.

These results show that the aggregated preference profile reflects the values of the sampled population.

\subsection{Finding 3: Question Framing Is a Silent Policy Lever}
\label{dis:gen:framing}

The third developer choice (how questions are framed) systematically shifts preferences and changes how moral foundations are associated with participants' judgments. In our study, framing reduced ideological differences in some cases but increased them in others. Prior research suggests that framing effects may occur without participants recognizing their influence~\cite{jakesch2023co}, which makes framing a silent policy lever rather than a neutral tool.
\smallskip

\noindent\textbf{Could-Be-You framing.}
Could-Be-You reduced several ideological gaps and was associated with longer response times in KIDNEY. Longer response times are consistent with greater processing effort, although they do not establish more reflective or deliberative reasoning. In KIDNEY, the framing reduced disagreement about vulnerable patients, including those with poor prognoses.

However, the same framing increased disagreement in GEN. Conservatives showed greater support for memorial requests than progressives. This pattern is consistent with perspective-taking increasing the salience of bereaved individuals' needs, although we did not directly measure empathy or perceived salience~\cite{rawls1971theory,bruno2024moral}. These results show that perspective-taking does not consistently produce convergence.
\smallskip

\noindent\textbf{World-You-Want framing.}
This framing also shifted preferences, but in different ways. In KIDNEY, it increased conservatives' preference for patients perceived as more responsible (e.g., those with healthier lifestyles)~\cite{sylvester2025s}. In WORK, it emphasized loyalty, leading conservatives to prioritize in-group protection over honesty.

These patterns suggest that this framing is not neutral. It can reinforce existing biases and political identities~\cite{jaques2025moral, mittal2023moral}. Response times were shorter than under Could-Be-You, but this difference does not identify the depth or quality of participants' reasoning. Overall, framing cannot be assumed to produce consensus.

\subsection{A Sensitivity Audit for Voting-Based Alignment}
\label{dis:gen:design}

Prior work sets out broad criteria for evaluating participatory machine learning systems, including stakeholder representation, elicitation, conflict resolution, and evaluation ~\cite{Feffer_ParticipatoryML_2023}. Our findings do not replace these established criteria. Instead, they adapt part of this framework to voting-based moral alignment by defining pipeline-specific sensitivity tests. 

\begin{figure}[t]
    \centering
    \includegraphics[width=\columnwidth]{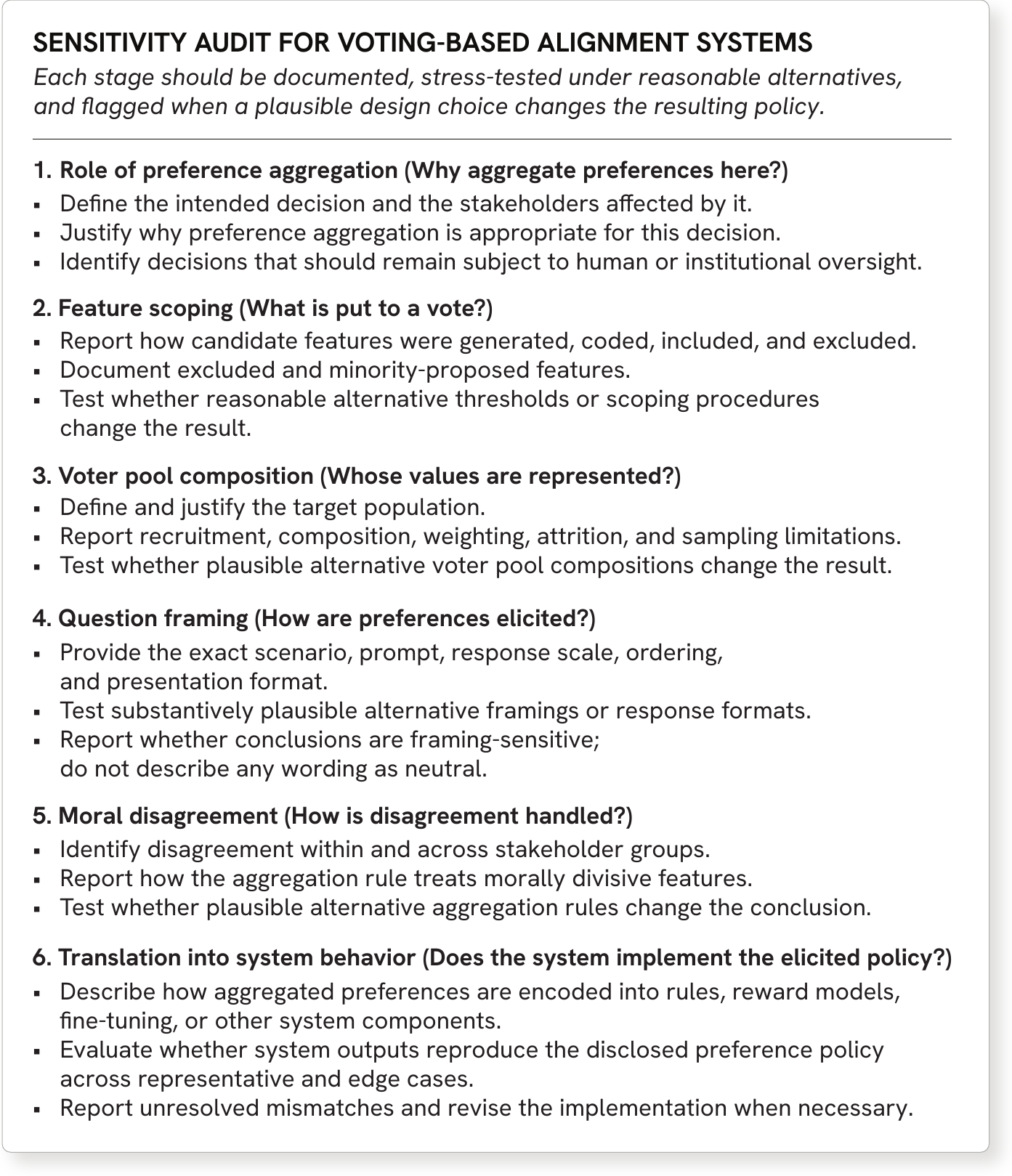}
    \caption{\textbf{A proposed sensitivity audit for voting-based alignment systems.} The checklist identifies pipeline choices that developers should document and test under plausible alternatives. The accompanying text defines consequential changes, and explains how developers should report and address them.}
    \label{fig:checklist}
\end{figure}

Figure~\ref{fig:checklist} presents an initial, empirically grounded sensitivity audit of the elicitation pipeline rather than a general checklist for participation. The recommendations on feature scoping, voter sampling, and question framing draw directly on our results. The recommendations on the role of aggregation, moral disagreement, and system implementation draw on the broader participatory machine learning and alignment literature~\cite{gabriel2020artificial,Feffer_ParticipatoryML_2023, participatoryML2024, ZhiXuan2024,Kneer2025hard, Baum2025}.

Our findings motivate specific sensitivity tests at each empirical stage. Differences in feature relevance across contexts show why developers should document and stress-test feature scopes. Ideology-linked differences and direction reversals show why they should test alternative voter pool compositions. Framing effects show why they should test alternative question formulations. Together, these checks show how strongly elicited preferences depend on pipeline choices before those preferences are used as the basis for a public or
collective policy.
\smallskip

\noindent\textbf{Recommendation 1: Define and justify the role of preference aggregation.}
Developers should state which decision the system will make or inform, which stakeholders it will affect, and why preference aggregation is appropriate for that decision. They should document intended uses,
foreseeable unintended uses and harms, and the authority retained by developers or deployers. They should also identify cases in which an aggregated preference should not determine system behavior. When the legitimacy of aggregation is contested, developers should supplement voting with stakeholder deliberation or other forms of institutional and normative oversight~\cite{Feffer_ParticipatoryML_2023}.
\smallskip

\noindent\textbf{Recommendation 2: Make feature scoping auditable.}

\noindent Developers should report how they generated and coded candidate features, the complete candidate set, the inclusion and exclusion criteria, the thresholds applied, and the features excluded from the final elicitation. They should also test alternative thresholds or scoping procedures. If an alternative adds or removes a feature, reverses the direction of a preference, or changes its relevance or disagreement classification, developers should report the result as sensitive to feature scope. They should then justify the selected scope and explain how they treated rare or minority-proposed features.
\smallskip

\noindent\textbf{Recommendation 3: Audit voter pool composition.}

\noindent Developers should define the target population and explain why its members should be represented. They should report recruitment procedures, demographic and ideological composition, exclusion and
attrition patterns, sampling limits, and any weighting rules~\cite{kallina2025stakeholder,vereschak2024trust,zhang2024empowering}. They should report group-specific results alongside the aggregate and recalculate estimates under plausible alternative voter pool compositions. If the direction or classification of a preference changes, or if its magnitude changes by more than the specified threshold, developers should describe the result as constituency-dependent rather than as a public preference.
\smallskip

\noindent\textbf{Recommendation 4: Audit question framing.}
Developers should disclose the complete scenario, exact prompt wording, response options, presentation order, and wording of any baseline condition. They should test plausible alternative framings or response formats and define consequential change in advance. If an alternative framing changes the direction or classification of a feature, or produces a change that exceeds the specified threshold, developers should report the result as framing-sensitive. They should justify the selected wording and should not describe any elicitation condition as normatively neutral
~\cite{jakesch2023co,rader2018explanations}.
\smallskip

\noindent\textbf{Recommendation 5: Preserve and document moral disagreement.}
Developers should report full response distributions and identify
features whose relevance or direction varies across groups~\cite{keswani2025can} or within individuals
~\cite{boerstler2024stability}. They should explain how the aggregation rule treats such disagreement and test whether plausible alternative rules produce different conclusions. If averaging conceals substantial disagreement or the resulting policy depends on the selected rule, developers should preserve disaggregated results. They should justify any decision to override, average, or otherwise resolve disagreement. Morally divisive cases may require deliberation or independent normative constraints rather than simple majority aggregation
~\cite{noothigattu2018voting,Feffer_ParticipatoryML_2023}.

\smallskip
\noindent\textbf{Recommendation 6: Validate the translation into system
behavior.}
Developers should describe how they translate aggregated preferences into rules, reward models, fine-tuning objectives, or other system components. They should test whether system outputs follow the disclosed preference policy across representative cases, edge cases, and morally divisive cases. A systematic mismatch between elicited preferences and system behavior should lead developers to revise the implementation or disclose the unresolved limitation. Because our study examined elicitation rather than model deployment, this recommendation draws on the broader literature and requires further empirical validation.

\subsection{Residual Limitations}
\label{dis:gen:ethical}

Our audit recommendations are necessary but not sufficient. Even with full transparency, aggregation remains a normative choice that cannot resolve deep value conflict.

\smallskip
\noindent\textbf{Pluralism and competing values.}
Our results show that different groups prioritize different moral values. Framing can widen these differences. A single aggregated policy therefore risks privileging one set of values. Alignment in pluralistic settings requires explicit trade-offs and transparency, not simple averaging.

\smallskip
\noindent\textbf{No universally optimal framing.}
Could-Be-You often promotes concern for the least advantaged, while World-You-Want can reinforce existing biases. However, neither framing works in all cases. When the least advantaged group is contested, perspective-taking can increase disagreement. No framing reliably produces consensus.

\smallskip
\noindent\textbf{Limits of majority aggregation.}
Even with full disclosure, aggregation risks a ``tyranny of the majority''~\cite{feffer2023moral}.  Public judgments are shaped by framing and identity~\cite{rehren2021moral, graham2009liberals}, and aggregation can blur rather than resolve value conflicts. Safeguards are therefore necessary to protect minority perspectives~\cite{tanksley2025ethics}. Voting-based alignment cannot, on its own, deliver fair or transparent outcomes.

\subsection{Limitations and Future Work}

Our study comes with six limitations. First, Likert scales measure importance rather than forced-choice decisions; future work should validate these findings using pairwise methods. Second, our design does not capture participants' reasoning; qualitative approaches, such as think-aloud protocols, could address this gap. Third, the feature inclusion threshold in Phase 1 is a design choice that may exclude minority views; future work should explore alternative thresholds. Fourth, we use AI literacy as a proxy for experience. Fifth, our use cases span different harm types and frequency but do not exhaust the space of applications. Sixth, our US-based sample limits generalizability across cultural contexts. Although our models account for several demographic characteristics of the sample, demographic and intersectional differences were beyond the scope of this study; future work should examine these differences in more diverse populations and non-Western cultural contexts \cite{zoshak2021beyond,atari2023morality}.
\section{Conclusion}

We examined three points at which upstream design choices shape moral preference elicitation: feature scoping, voter sampling, and question framing. Across three deployment contexts, the analyses show that elicited preferences are contingent on how each stage is configured. First, \textit{feature scoping} is context-dependent. The features participants consider morally relevant differ across use cases, which limits transfer across domains. Second, \textit{voter sampling} shapes outcomes. Differences in political ideology lead to systematic shifts in judgments and, in some cases, reverse the direction of preferences. Third, \textit{question framing} alters judgments and the estimated relationships between moral foundations and those judgments. Framing can reduce some disagreements but amplify others.

These results show that moral preference elicitation does not produce a single, stable set of values. Instead, the output depends on upstream design choices made by developers. What appears as ``public morality'' in a deployed system is, in part, a function of how the pipeline is constructed.

The main takeaway is that voting-based alignment does not remove human judgment from AI systems; it relocates it to design choices about features, samples, and framing. Systems that rely on preference aggregation should treat elicited preferences as contingent rather than fixed. This requires disclosing how features are defined, participants sampled, and questions framed, and assessing how sensitive outputs are to each. Without such disclosure, claims of neutrality or fairness are difficult to evaluate.

Our findings suggest that alignment cannot rely on aggregation alone. If preferences shift with context, population, and framing, then alignment also requires normative judgment and institutional design. This may include structured deliberation, stakeholder representation, and mechanisms for accountability. Future research should extend this analysis to additional domains, test alternative elicitation methods that reduce sensitivity to framing and sampling, and explore how to integrate empirical preferences with normative frameworks in system design.
\section*{Researcher Positionality Statement}

Our team consists of three members (two men, one woman) from East Asia, Southern Europe, and Eastern Europe, now based across Europe and North America. We bring diverse ethnic, religious, and cultural backgrounds and expertise spanning social and cognitive psychology, Responsible AI, HCI, Computer Science, and NLP across academic and industry research settings. Having lived across different political systems, we acknowledge that our positionality may have influenced various aspects of our research, including our choice of use cases, design of question framing interventions, and interpretation of results across political groups. We recognize the importance of including a broader range of voices from academia, industry, and underrepresented regions and communities.
\bigskip

\section*{Ethical Statement}
Our work raises four ethical considerations. First, the study was conducted with our organization's approval. We adhered to established guidelines for human subjects research, ensuring that no personal identifiers were collected, personal information was removed from all data, and access was restricted to the research team. Participants were recruited via Prolific, compensated at a rate of at least 8 USD/hour, and free to withdraw at any time.

Second, two of the three use cases involved potentially sensitive topics: life-or-death kidney allocation and generative AI depictions of deceased individuals. To mitigate distress, we consulted two domain experts and presented the use cases in standardized, non-evaluative wording. We do not regard the resulting descriptions as normatively neutral across conditions. The Control condition serves only as a baseline with no additional question framing prompt.

Third, participants were sampled to balance political ideology across three categories: conservative, moderate, and progressive. We acknowledge that this categorization is a simplification that may not capture the full diversity of political views, particularly non-Western ones.

Fourth, this work identifies how feature scoping, voter sampling, and question framing can shape AI policy. While our intent is to promote transparency and accountability, the same findings could inform manipulation of elicitation pipelines. We therefore frame our contributions as a sensitivity audit and recommendations for voting-based alignment rather than as prescriptive design rules.

\section*{Generative AI Usage}
The authors used \texttt{GPT-4o-mini} and \texttt{Gemini 2.0} during the preparation of this manuscript. These tools were employed to assist with grammar and style editing, text summarization, and the structuring of figures and tables. Additionally, these tools supported the development of computer code used in the research and provided assistance in the content analysis of user responses. Large Language Models were not used to generate original publication text. All final text, interpretations, and conclusions were authored and verified by the human researchers to ensure originality and integrity. 

\bibliography{main} 

%%%%%% APPENDIX %%%%%% 
\appendix
\onecolumn
\makeatletter
\def\section{\@startsection {section}{1}{\z@}{-2.0ex plus
-0.5ex minus -.2ex}{3pt plus 2pt minus 1pt}{\Large\bf\raggedright}}
\makeatother

\begin{center}
    {\LARGE\bfseries Participatory Moral AI Is Not Neutral:\\ The Invisible Hand of Developers\par}
    \bigskip
    {\large\bfseries Taenyun Kim\textsuperscript{1}, Edyta Bogucka\textsuperscript{2}, Daniele Quercia\textsuperscript{2,3}\par}
    \medskip
    {\large
    \textsuperscript{1}Michigan State University, US\\
    \textsuperscript{2}Nokia Bell Labs, Cambridge, UK\\
    \textsuperscript{3}Politecnico di Torino, Italy\\
    kimtaeny@msu.edu, edyta.bogucka@nokia-bell-labs.com, daniele.quercia@nokia-bell-labs.com
    \par}
    \bigskip
    \bigskip
    {\Large\bfseries Supplementary Materials\par}
\end{center}

\bigskip

\section{Demographic Characteristics of Phase 1 and Phase 2 Study Participants}
\label{app:demographics}
\begin{table}[h]
\centering

\caption{Demographic characteristics of Phase 1 participants across three use cases, including political ideology, age, gender, race/ethnicity, and education level. Percentages are reported with sample counts in parentheses; age is reported as mean $\pm$ \textit{SD}.}

\label{tab:demographics_s1}
\begin{tabular}{llccc}
\toprule
 & & \makecell[t]{\textbf{Use case 1: KIDNEY}\\AI kidney allocation}&  \makecell[t]{\textbf{Use case 2: WORK}\\AI agents simulating\\absent workers} &  \makecell[t]{\textbf{Use case 3: GEN}\\Generative AI content \\ of the deceased} \\
\midrule
Sample size & $N$ & 150 & 149 & 150 \\
\midrule
\multirow{3}{*}{Political ideology} 
 & Conservative & 33.3\% (n=50) & 32.9\% (n=49) & 33.3\% (n=50) \\
 & Moderate     & 33.3\% (n=50) & 33.6\% (n=50) & 33.3\% (n=50) \\
 & Progressive  & 33.3\% (n=50) & 33.6\% (n=50) & 33.3\% (n=50) \\
\midrule
Age & Mean (SD) & 44.63 (13.34) & 47.87 (13.33) & 44.75 (14.12) \\
\midrule
\multirow{2}{*}{Gender} 
 & Female & 50.0\% (n=75) & 49.7\% (n=74) & 50.0\% (n=75) \\
 & Male   & 50.0\% (n=75) & 50.3\% (n=75) & 50.0\% (n=75) \\
\midrule
\multirow{5}{*}{Race/Ethnicity} 
 & White  & 86.7\% (n=130) & 83.9\% (n=125) & 86.7\% (n=130) \\
 & Black  & 4.0\% (n=6)    & 7.4\% (n=11)   & 4.0\% (n=6) \\
 & Asian  & 5.3\% (n=8)    & 2.7\% (n=4)    & 4.7\% (n=7) \\
 & Mixed  & 2.0\% (n=3)    & 4.0\% (n=6)    & 3.3\% (n=5) \\
 & Other  & 2.0\% (n=3)    & 2.0\% (n=3)    & 1.3\% (n=2) \\
\midrule
\multirow{3}{*}{Education} 
 & Pre-college       & 20.0\% (n=30) & 22.1\% (n=33) & 21.3\% (n=32) \\
 & College degree    & 58.7\% (n=88) & 49.7\% (n=74) & 50.7\% (n=76) \\
 & Advanced degree   & 21.3\% (n=32) & 28.2\% (n=42) & 28.0\% (n=42) \\
\bottomrule
\end{tabular}
\end{table}

\newpage

\begin{table}[t]
\centering
\caption{Demographic characteristics of Phase 2 participants across three use cases, including political ideology, age, gender, race/ethnicity, education level, AI literacy, religiosity, and area of residence. Percentages are reported with sample counts in parentheses; age and AI literacy are reported as mean $\pm$ \textit{SD}.}

\label{tab:demographics_s2}
\begin{tabular}{llccc}
\toprule
 & & \makecell[t]{\textbf{Use case 1: KIDNEY}\\AI kidney allocation}&  \makecell[t]{\textbf{Use case 2: WORK}\\AI agents simulating\\absent workers} &  \makecell[t]{\textbf{Use case 3: GEN}\\Generative AI content \\ of the deceased} \\
\midrule
Sample size & $N$ & 120 & 120 & 120 \\
\midrule
\multirow{2}{*}{Political ideology} 
 & Conservative & 50.0\% (n=60) & 50.0\% (n=60) & 50.0\% (n=60) \\
 & Progressive      & 50.0\% (n=60) & 50.0\% (n=60) & 50.0\% (n=60) \\
\midrule
Age & Mean (SD) & 47.30 (16.54) & 47.85 (15.59) & 48.44 (16.62) \\
\midrule
\multirow{2}{*}{Gender} 
 & Female & 50.0\% (n=60) & 50.0\% (n=60) & 50.0\% (n=60) \\
 & Male   & 50.0\% (n=60) & 50.0\% (n=60) & 50.0\% (n=60) \\
\midrule
\multirow{6}{*}{Race/Ethnicity} 
 & White (non-Hispanic) & 69.2\% (n=83) & 69.2\% (n=83) & 66.7\% (n=80) \\
 & Black                & 21.7\% (n=26) & 16.7\% (n=20) & 19.2\% (n=23) \\
 & Asian                & 0.8\% (n=1)   & 0.8\% (n=1)   & 0.0\% (n=0) \\
 & Mixed                & 2.5\% (n=3)   & 6.7\% (n=8)   & 7.5\% (n=9) \\
 & Other                & 2.5\% (n=3)   & 1.7\% (n=2)   & 1.7\% (n=2) \\
 & Hispanic             & 3.3\% (n=4)   & 5.0\% (n=6)   & 5.0\% (n=6) \\
\midrule
\multirow{3}{*}{Education} 
 & Pre-college       & 20.0\% (n=24) & 20.0\% (n=24) & 25.8\% (n=31) \\
 & College degree    & 49.2\% (n=59) & 50.8\% (n=61) & 47.5\% (n=57) \\
 & Advanced degree   & 30.8\% (n=37) & 29.2\% (n=35) & 26.7\% (n=32) \\
\midrule
AI literacy & Limitations/Ethics (M, SD) & 0.76 (0.22) & 0.80 (0.20) & 0.79 (0.21) \\
            & Technical (M, SD)          & 0.58 (0.24) & 0.62 (0.23) & 0.60 (0.23) \\
\midrule
\multirow{5}{*}{Religiosity} 
 & Not at all important  & 30.8\% (n=37) & 30.8\% (n=37) & 33.3\% (n=40) \\
 & Slightly important    & 10.0\% (n=12) & 10.8\% (n=13) & 11.7\% (n=14) \\
 & Moderately important  & 14.2\% (n=17) & 10.8\% (n=13) & 12.5\% (n=15) \\
 & Very important        & 30.0\% (n=36) & 22.5\% (n=27) & 24.2\% (n=29) \\
 & Extremely important   & 15.0\% (n=18) & 25.0\% (n=30) & 18.3\% (n=22) \\
\midrule
\multirow{2}{*}{Area of residence} 
 & Rural & 30.0\% (n=36) & 30.8\% (n=37) & 25.8\% (n=31) \\
 & Urban & 70.0\% (n=84) & 69.2\% (n=83) & 74.2\% (n=89) \\
\bottomrule
\end{tabular}
\end{table}
\clearpage

\section{Instructions Provided to Participants in Phase 1 and Phase 2}
\label{app:phase_instructions}
\subsection{Phase 1 Instructions: Identifying Morally Relevant and Irrelevant Features}
\label{inst:study1}

\begin{figure*}[ht!]
	\centering
	\includegraphics[width=0.61\textwidth]{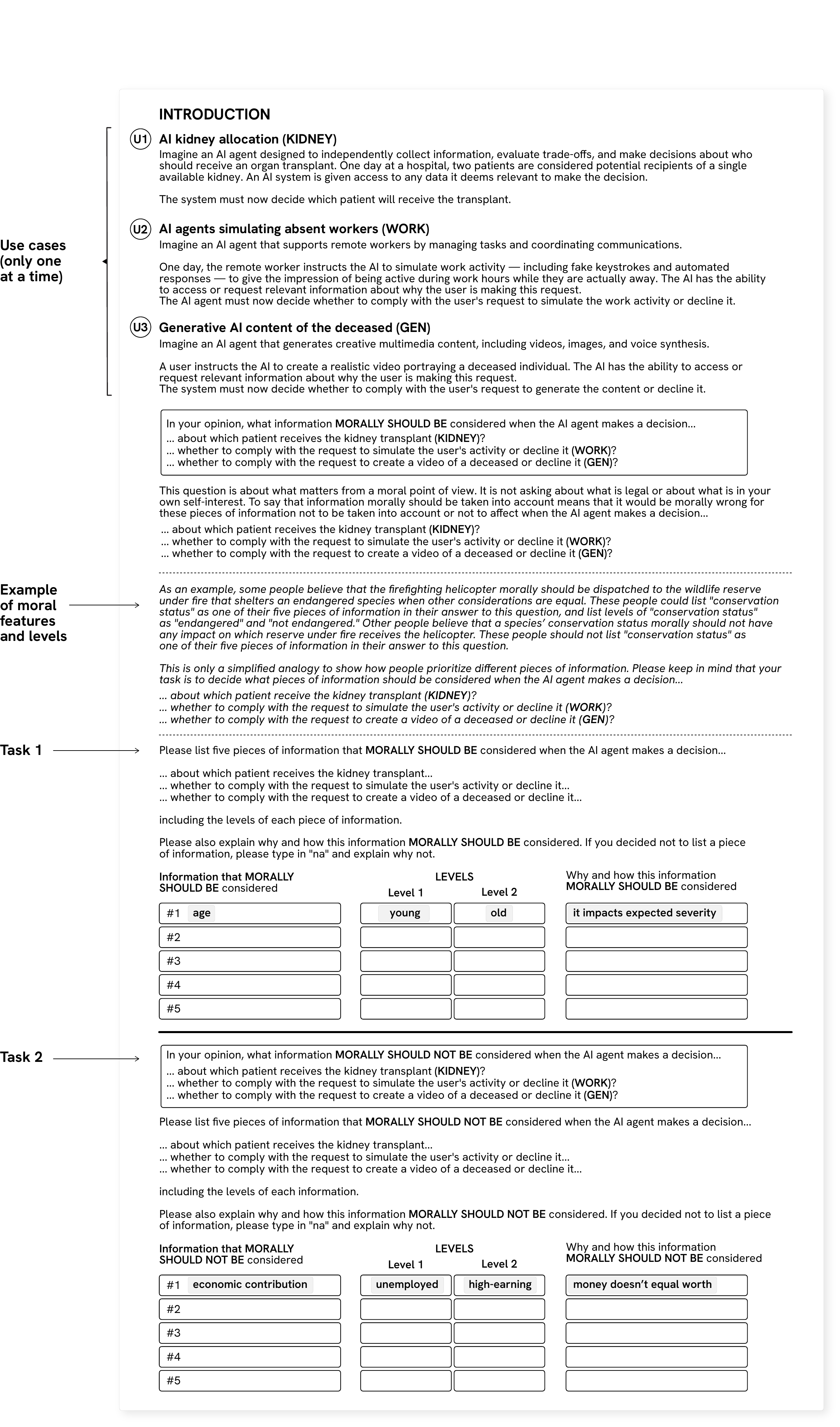}
    \caption{\textbf{Example of the Phase 1 interface.} Participants were randomly assigned to one of three use cases: AI kidney allocation (KIDNEY), AI agents simulating absent workers (WORK), or generative AI content of the deceased (GEN). They first read an introduction to their assigned use case. They then read an illustrative example, unrelated to any of the three use cases, that clarified what counted as a moral feature and its levels. They then completed two tasks. In Task 1, they listed five pieces of information they believed the AI agent morally should consider, along with two levels for each and a free-text justification. In Task 2, they completed the same procedure for information they believed the AI agent morally should not consider.}
	\label{app:instructions}
\end{figure*}

\subsection{Phase 2 Instructions: Evaluating the Moral Weight of Features Under Three Framing Conditions}
\label{inst:study2}

\begin{figure*}[ht!]
	\centering
	\includegraphics[width=0.8\textwidth]{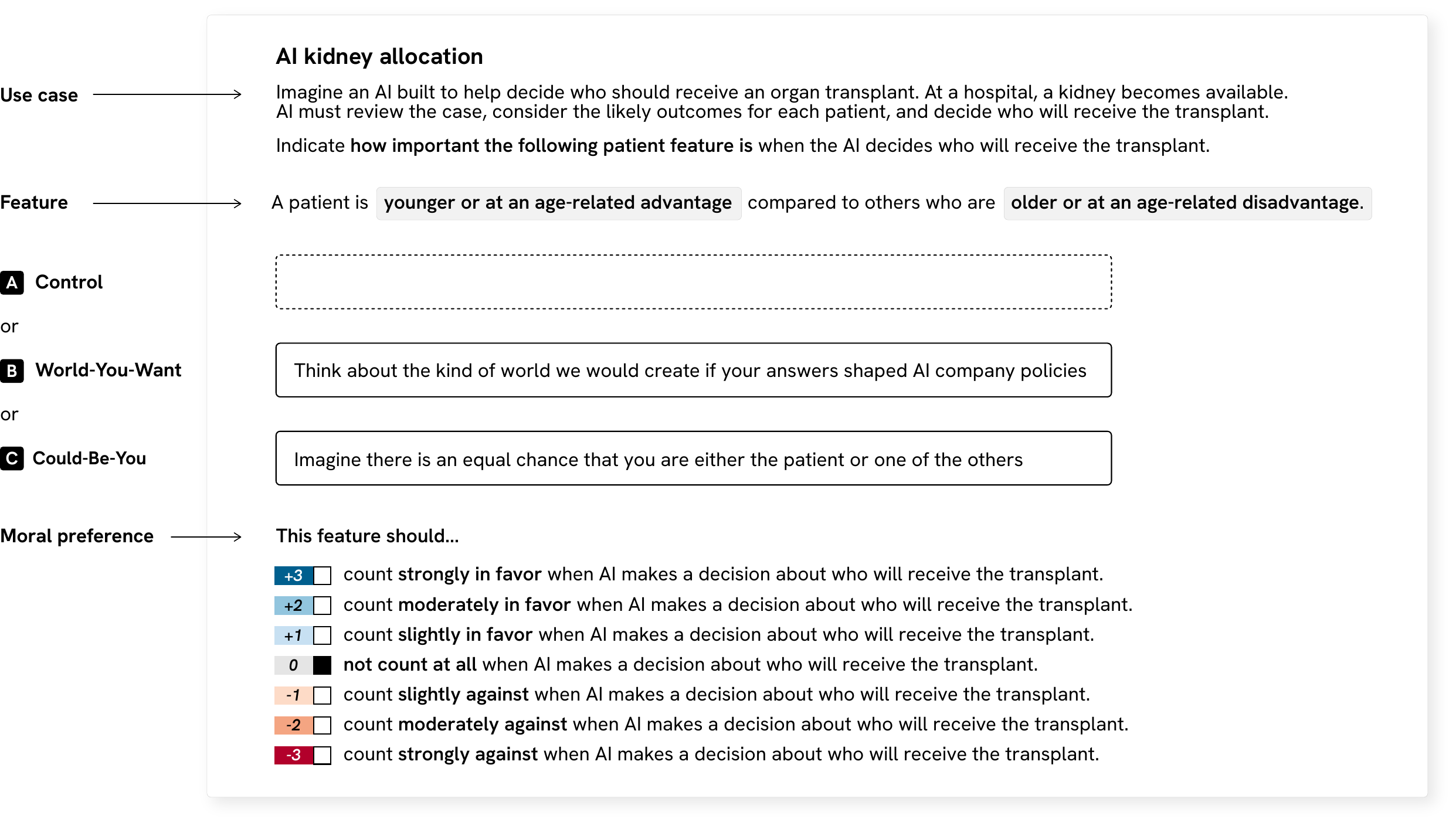}
    \caption{\textbf{Example of the Phase 2 interface for the AI kidney allocation use case (KIDNEY).} Participants were shown a feature framed as a contrast (here: age presented as a contrast between younger \emph{vs.} older patients), and asked to rate its moral importance on a 7-point scale from -3 (strongly against) to +3 (strongly in favor). Depending on condition, participants received additional framing instructions: \textit{Control} (A), \textit{World-You-Want} (B), or \textit{Could-Be-You} (C).}
	\label{app:study2_kidney_treatment_appendix}
\end{figure*}

\begin{figure*}[ht!]
	\centering
	\includegraphics[width=0.8\textwidth]{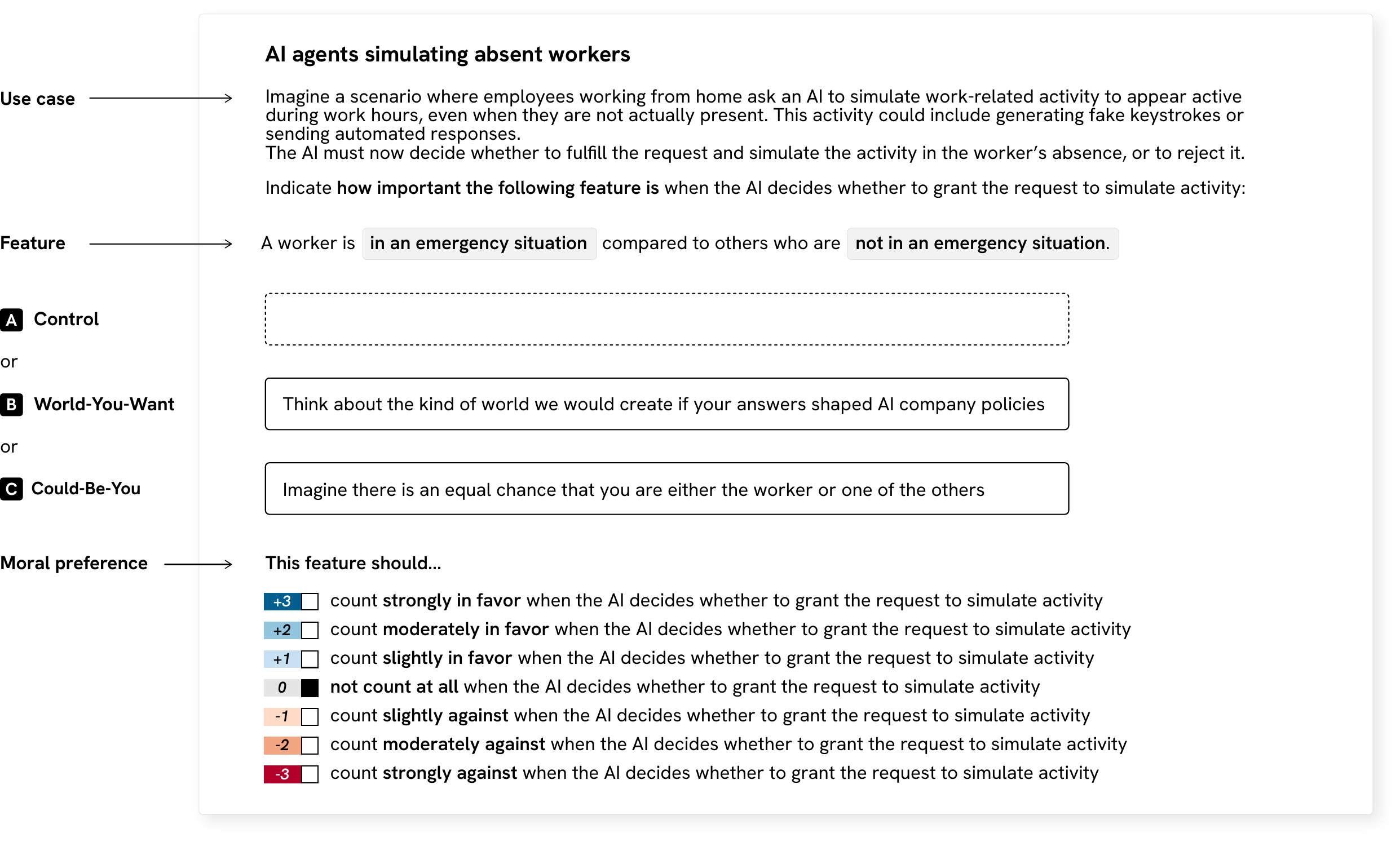}
    \caption{\textbf{Example of the Phase 2 interface for the AI agents simulating absent workers use case (WORK).} Participants were shown a feature framed as a contrast (here: a personal emergency situation, presented as a contrast between a worker in an emergency situation \emph{vs.} others who are not in an emergency situation), and were asked to rate its moral importance on a 7-point scale from -3 (strongly against) to +3 (strongly in favor). Depending on condition, participants received additional framing instructions: \textit{Control} (A), \textit{World-You-Want} (B), or \textit{Could-Be-You} (C).}
	\label{app:study2_work}
\end{figure*}

\begin{figure*}[ht!]
	\centering
	\includegraphics[width=0.8\textwidth]{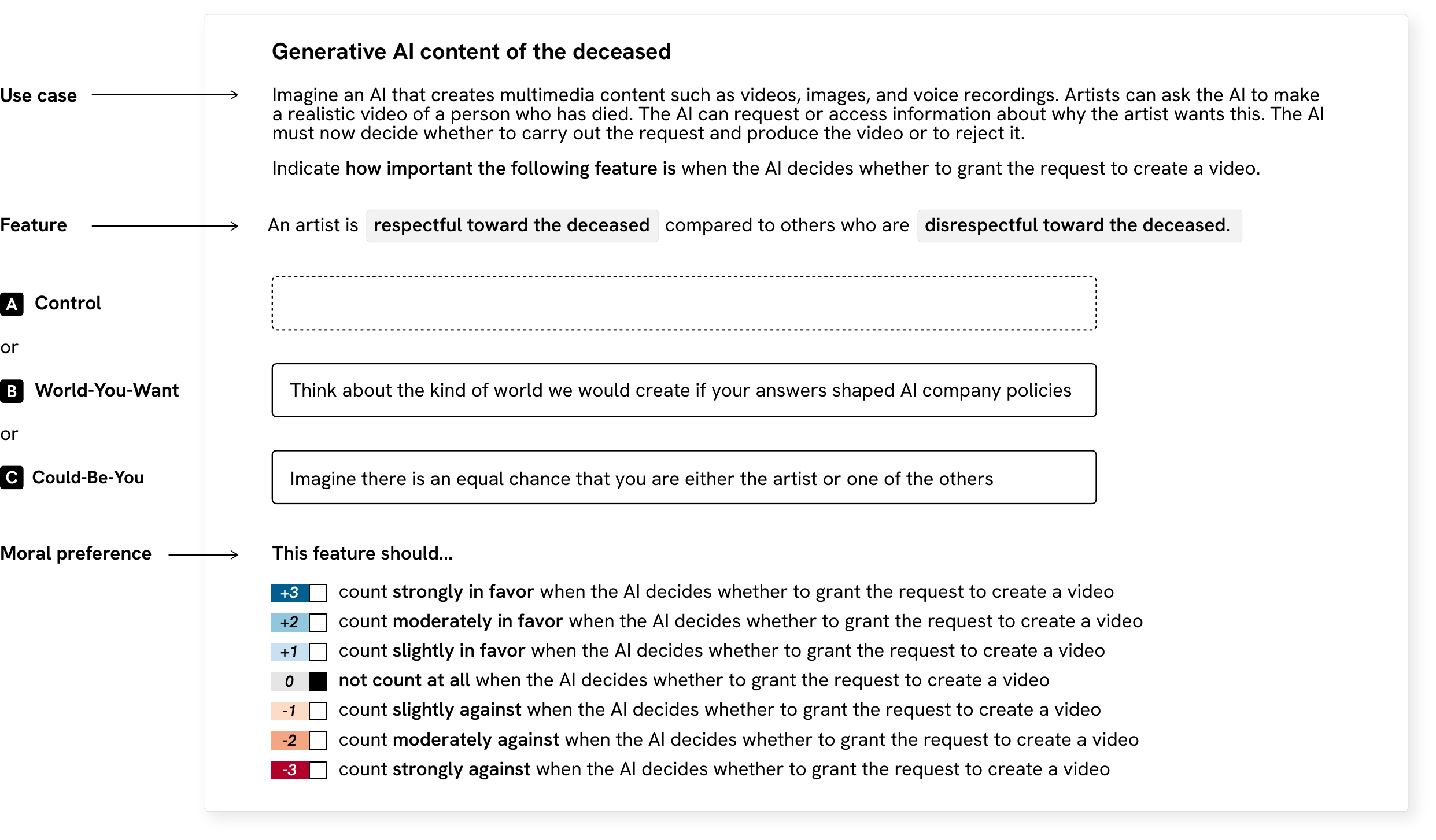}
    \caption{\textbf{Example of the Phase 2 interface for the generative AI content of the deceased use case (GEN).} Participants were shown a feature framed as a contrast (here: respect toward the deceased, presented as a contrast between being respectful \emph{vs.} disrespectful toward the deceased), and were asked to rate its moral importance on a 7-point scale from -3 (strongly against) to +3 (strongly in favor). Depending on condition, participants received additional framing instructions: \textit{Control} (A), \textit{World-You-Want} (B), or \textit{Could-Be-You} (C).}
	\label{app:study2_art}
\end{figure*}
\clearpage

\section{Language Model Prompt for Identifying Moral Features from Participant Responses (Phase 1)}
\label{app:llm-prompt}
\begin{tcolorbox}[
    colback=white,
    colframe=black,
    width=\textwidth,
    arc=1mm,
    boxrule=0.25mm,
    fontupper=\footnotesize
]
\medskip
\textbf{System role:} You are an expert in HCI, specialized in qualitative methodologies. You are analyzing responses from participants asked: ``Which features SHOULD or SHOULD NOT be morally considered when an AI agent makes a decision?''
\newline

\textbf{User role:} You are analyzing responses from participants asked to imagine this use case (one from the list below): 
\newline

\textbf{Use case 1: KIDNEY} Imagine an AI agent designed to independently collect information, evaluate trade-offs, and make decisions about who should receive an organ transplant. One day at a hospital, two patients are considered potential recipients of a single available kidney. An AI system is given access to any data it deems relevant to make the decision. The system must now decide which patient will receive the transplant.

\textbf{Use case 2: WORK} Imagine an AI agent that supports remote workers by managing tasks and coordinating communications. One day, the remote worker instructs the AI to simulate work activity — including fake keystrokes and automated responses — to give the impression of being active during work hours while they are actually away. The AI has the ability to access or request relevant information about why the user is making this request. The AI agent must now decide whether to comply with the user's request to simulate the work activity or decline it.

\textbf{Use case 3: GEN} Imagine an AI agent that generates creative multimedia content, including videos, images, and voice synthesis. A user instructs the AI to create a realistic video portraying a deceased individual. The AI has the ability to access or request relevant information about why the user is making this request. The system must now decide whether to comply with the user’s request to generate the content or decline it.
\newline

The file contains the following columns:

- SID\textunderscore anonymized: response id

- info: the participant's mention of features they believe are morally relevant or irrelevant for the scenario

- moral\textunderscore consider: indication if the features should or should not be morally considered for the scenario

- why: the participant's brief explanation of why these features should or should not be morally considered

- level1, level2: two specific values (or ``levels'' or ``states'') that are relevant in the moral reasoning context for the features that should be considered
\newline

\textbf{TASK:} Analyze the responses and return a structured annotation following these four clearly labeled steps.
\newline

\textbf{Step 1:} Identify specific features explicitly mentioned or implied by the participant. Always map these to a single, standardized feature name; never use synonyms or participant-specific phrasing. Infer implied features where necessary (e.g., ''someone always reliable'' implies ''past reliability''). Only include features that could plausibly appear in a real-world AI decision system. Avoid vague or abstract concepts.

Examples of valid features: ``past productivity'', ``health'', ``personal emergency'', ``number of dependents''

Examples of invalid features: ``their life'', ``everything about them''
\newline

\textbf{Step 2:} For each feature identified in Step 1, determine at least two of its specific values ( ``levels'' or ``states''):

If values are present in the level1 and level2 columns:

- Use these as the explicit values for that feature.

- If level1 or level2 contain ``na'', ``nan'', or are blank/empty, do not include these as values for the feature.

- Mark these as explicit.

If level1 and level2 are not specified or contain ``na''/``nan'':

- Carefully infer plausible, concrete values for the feature based on the info and why columns.

- Use only information that is reasonable given the participant's response and real-world context.

- Use natural language phrases, not generic labels like ``low'' or ``high'' unless that's what the participant implies.

- Mark these values as inferred.

For each feature, values must fill this template exactly: [patient / worker / artist] is [value\textunderscore 1] compared to others who are [value\textunderscore 2]
\newline

General Rules for Step 2:

- Always strictly follow the template.

- Only include plausible, real-world features and values (e.g., ``chronic illness'', ``excellent performance'', ``unavoidable emergency'') consistent with how the feature might appear in real-world AI decision-making.

- If uncertain, err on the side of specificity and real-world plausibility.

- Do not invent or generalize features or values beyond what is supported by the participant's response and context.
\newline

\noindent\textit{Prompt continued on next page}
\medskip
\end{tcolorbox}

\begin{tcolorbox}[
    colback=white,
    colframe=black,
    width=\textwidth,
    arc=1mm,
    boxrule=0.25mm,
    fontupper=\footnotesize
]

\medskip
\noindent\textit{Prompt continued from previous page}
\newline

\textbf{Step 3:} Provide concise summary of the participant's moral reasoning. Avoid generic or one-word summaries like ``fairness'' unless no more detail is available.

Examples of valid summaries: ``fairness due to factors beyond control'', ``privacy concerns'', ``irrelevance to job performance'', ``accountability for choices''
\newline

\textbf{Step 4:} Determine moral relevance. Label whether the feature was seen as:

- ``should'' (morally relevant)

- ``should not'' (morally irrelevant)

- ``mixed'' (multiple features with differing views)

- ``unclear'' (if input is empty, incoherent or vague)
\newline

Here are the RESPONSES: 
SID\textunderscore anonymized: ``{}'', 

info: ``{}'',

moral\textunderscore consider: ``{}'',

why: ``{}'',

level1: ``{}'',

level2: ``{}''
\newline

\textbf{RESPONSE FORMAT:}

Return your analysis strictly in this JSON format:

\verb|{{|

        ``id'': ``SID\textunderscore anonymized'',
        
        ``info'': ``info'',
        
        ``features'': [``feature\textunderscore 1'', ``feature \textunderscore 2''],
        
        ``values\textunderscore per \textunderscore feature'': \verb|{{|
        
            ``feature\textunderscore 1'': [
            
            \verb|{{|``value'': ``value\textunderscore 1'', ''type'': ``explicit'' or ``inferred''\verb|}}|,
            
            \verb|{{|``value'': ``value\textunderscore 2'', ``type'': ``explicit'' or ``inferred''\verb|}}|,
            
            ],
            
            ``feature\textunderscore 2'': [
            
            \verb|{{|``value'': ``value\textunderscore 1'', ``type'': ``explicit'' or ``inferred''\verb|}}|,
            
            \verb|{{|``value'': ``value\textunderscore 2'', ``type'': ``explicit'' or ``inferred''\verb|}}|,
            
            ]
            
        \verb|}}|,
        
        ``templates\textunderscore per \textunderscore feature'' : {{
        
            ``feature\textunderscore 1'': ``A patient is [value\textunderscore 1] compared to others who are [value\textunderscore 2]'',
            
            ``feature\textunderscore 2'': ``A patient is [value\textunderscore 1] compared to others who are [value\textunderscore 2]'',
            
        }}
        ``justification'': ``short summary of the moral reasoning'',
        
        ``moral\textunderscore relevance'': ``should'' or ``should not'' or ``mixed'' or ``unclear''
        
\verb|}}|
\newline

JSON Formatting Rules: 

Rule 1: Go systematically through each of the ***responses***. 
       
Rule 2: Ensure that the JSON response strictly follows the format  

Rule 3: Output only JSON - do not include any other content. 
\medskip
        
\end{tcolorbox}

\clearpage

\section{Moderated Mediation Path Model: Political Ideology and Moral Foundations Moderated by Question Framing (Phase 2)}
\label{app:model}

\begin{figure*}[h]
    \centering
    \includegraphics[width=0.9\textwidth]{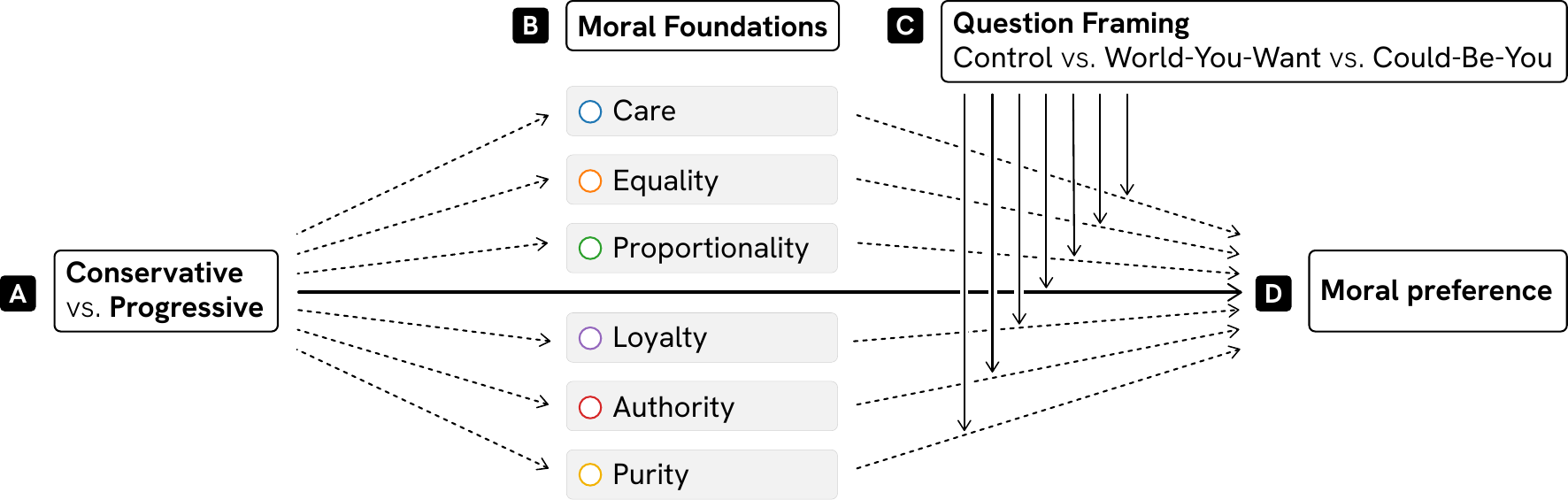}
    \caption{\textbf{Moderated mediation path model used in Phase 2 to examine the role of political ideology, moral foundations, and question framing on moral preferences}. The path model (Hayes's \citeyear{hayes2017introduction}, Model 15): \emph{(A)} Political ideology (conservative \emph{vs.} progressive) influences endorsement of \emph{(B)} six moral foundations (Care, Equality, Proportionality, Loyalty, Authority, and Purity), which in turn, being moderated by \emph{(C)} question framings, predict \emph{(D)} the perceived moral importance of features in AI decision-making. Question framing (Control \emph{vs.} World-You-Want \emph{vs.} Could-Be-You) moderates the relationship between political ideology and perceived moral importance by either attenuating or amplifying both the direct effect (\emph{A} $\rightarrow$ \emph{D}) and the indirect effect via moral foundations (\emph{A} $\rightarrow$ \emph{B} $\rightarrow$ \emph{D}). The covariates were sex (1 = female), age, education level (1 = pre-college; 2 = college degree; 3 = advanced degree), ethnicity (1 = non-Hispanic White), religiosity (1 = not at all important; 5 = extremely important in daily life), area of residence (1 = rural), and AI literacy regarding AI limitations, ethical considerations, and technical understanding.}
    \label{fig:study2_model}
\end{figure*}
\clearpage

\section{Phase 1 and Phase 2 Results for Use Case 1: AI Kidney Allocation (KIDNEY)}
\label{app:results_kidney}

\begin{table}[htbp]
    \centering
    \small
    \caption{Moral features identified in Phase 1 as \textit{should be considered} for the AI kidney allocation use case (KIDNEY), color-coded by mention rate: \textcolor{502}{\textbf{dark blue ($\ge$50\%)}}, \textcolor{252}{\textbf{medium blue} ($\ge$25\%)}, \textcolor{102}{\textbf{light blue ($\ge$10\%)}}, and \textcolor{02}{\textbf{gray ($<$10\%)}}. Features mentioned by fewer than 2.5\% of participants are omitted.}
    \label{tab:kidney-study1-should}
    \begin{tabular}{@{\extracolsep{\fill}} p{5.5cm} c c c c}
    \toprule
    \textbf{Moral features} & \textbf{Overall (\%)} & \multicolumn{3}{c}{\textbf{Political ideology(\%)}}\\
    \cmidrule(lr){3-5}
    \textbf{SHOULD BE CONSIDERED}& & \textbf{Progressive} & \textbf{Moderate} & \textbf{Conservative} \\
    \midrule
    \rowcolor{50}Age & 82 & 32 & 33 & 35 \\
    \hline
    \rowcolor{25}Health status & 47 & 40 & 30 & 30 \\
    \rowcolor{25}Chance of survival & 35 & 40 & 26 & 34 \\
    \rowcolor{25}Waiting time for transplant & 29 & 41 & 27 & 32 \\
    \hline
    \rowcolor{10}Urgency of transplant & 22 & 33 & 33 & 34 \\
    \rowcolor{10}Expected lifespan after transplant & 18 & 41 & 33 & 26 \\
    \rowcolor{10}Caregiving status & 17 & 32 & 32 & 36 \\
    \rowcolor{10}Additional illnesses & 16 & 25 & 50 & 25 \\
    \rowcolor{10}Lifestyle & 15 & 26 & 39 & 35 \\
    \rowcolor{10}Criminal history & 11 & 31 & 25 & 44 \\
    \hline
    \rowcolor{0}Adherence to aftercare & 9 & 31 & 23 & 46 \\
    \rowcolor{0}Severity of the condition & 9 & 15 & 38 & 47 \\
    \rowcolor{0}Family situation & 7 & 27 & 36 & 37 \\
    \rowcolor{0}Compatibility with donor organ & 6 & 33 & 33 & 34 \\
    \rowcolor{0}Smoking status & 6 & 56 & 11 & 33 \\
    \rowcolor{0}Alcohol use & 6 & 44 & 33 & 23 \\
    \rowcolor{0}Quality of life & 5 & 38 & 50 & 12 \\
    \rowcolor{0}Responsibility for illness & 5 & 43 & 14 & 43 \\
    \rowcolor{0}Weight & 4 & 17 & 33 & 50 \\
    \rowcolor{0}Organ damage level & 4 & 17 & 50 & 33 \\
    \rowcolor{0}Previous transplant history & 3 & 40 & 20 & 40 \\
    \rowcolor{0}Risk of organ acceptance & 3 & 0 & 80 & 20 \\
    \rowcolor{0}Drug use & 3 & 0 & 60 & 40 \\
    \rowcolor{0}Substance abuse & 3 & 25 & 50 & 25 \\
    \rowcolor{0}Ethnicity & 3 & 25 & 50 & 25 \\
    \rowcolor{0}Consent & 3 & 50 & 0 & 50 \\
    \rowcolor{0}Cause of illness & 3 & 25 & 25 & 50 \\
    \rowcolor{0}Religion & 3 & 25 & 25 & 50 \\
    \bottomrule
\end{tabular}
\end{table}

\newpage
\begin{table}[htbp]
    \centering
    \small
    \caption{Moral features identified in Phase 1 as \textit{should not be considered} for the AI kidney allocation use case (KIDNEY), color-coded by mention rate: \textcolor{502}{\textbf{dark blue ($\ge$50\%)}}, \textcolor{252}{\textbf{medium blue} ($\ge$25\%)}, \textcolor{102}{\textbf{light blue ($\ge$10\%)}}, and \textcolor{02}{\textbf{gray ($<$10\%)}}. Features mentioned by fewer than 2.5\% of participants are omitted.}
    \label{tab:kidney-study1-shouldnt}
    \begin{tabular}{@{\extracolsep{\fill}} p{5.5cm} c c c c}
    \toprule
    \textbf{Moral features} & \textbf{Overall (\%)} & \multicolumn{3}{c}{\textbf{Political ideology(\%)}}\\
    \cmidrule(lr){3-5}
    \textbf{SHOULD NOT BE CONSIDERED}& & \textbf{Progressive} & \textbf{Moderate} & \textbf{Conservative} \\
    \midrule
    \rowcolor{50}Ethnicity & 66 & 34 & 29 & 37 \\
    \rowcolor{50}Gender & 64 & 35 & 29 & 36 \\
    \rowcolor{50}Wealth & 52 & 32 & 31 & 37 \\
    \hline
    \rowcolor{25}Religion & 40 & 30 & 23 & 47 \\
    \rowcolor{25}Social status & 27 & 32 & 29 & 39 \\
    \hline
    \rowcolor{10}Sexual orientation & 22 & 33 & 33 & 34 \\
    \rowcolor{10}Occupation & 12 & 44 & 22 & 34 \\
    \rowcolor{10}Age & 11 & 29 & 59 & 12 \\
    \rowcolor{10}Geographic location & 11 & 35 & 29 & 36 \\
    \rowcolor{10}Political ideology & 10 & 27 & 33 & 40 \\
    \hline
    \rowcolor{0}Nationality & 9 & 21 & 57 & 22 \\
    \rowcolor{0}Family situation & 7 & 40 & 30 & 30 \\
    \rowcolor{0}Employment status & 7 & 30 & 20 & 50 \\
    \rowcolor{0}Disability status & 7 & 40 & 50 & 10 \\
    \rowcolor{0}Criminal history & 6 & 33 & 44 & 23 \\
    \rowcolor{0}Caregiving status & 6 & 44 & 44 & 12 \\
    \rowcolor{0}Popularity & 4 & 17 & 50 & 33 \\
    \rowcolor{0}Marital status & 3 & 60 & 0 & 40 \\
    \rowcolor{0}Economic status & 3 & 40 & 40 & 20 \\
    \rowcolor{0}Education & 3 & 40 & 40 & 20 \\
    \rowcolor{0}Mental health problems & 3 & 25 & 50 & 25 \\
    \rowcolor{0}Appearance & 3 & 25 & 25 & 50 \\
    \rowcolor{0}Past behaviour & 3 & 0 & 25 & 75 \\
    \bottomrule
\end{tabular}
\end{table}
\clearpage
\begin{table}[t]
    \caption{Moral relevance and importance of features elicited in Phase 2 for the AI kidney allocation use case (KIDNEY). \textit{Note:} $^{***}p<0.001$, $^{**}p<0.01$, $^{*}p<0.05$. Significance levels are based on two separate two-sided one-sample \textit{t}-tests: Mean (Coded) was tested against a reference value of 0.5; Mean was tested against 0.}
	\small
	\label{tab:kidney_study2}
	\begin{tabular}{llll}
		\toprule
		\textbf{Feature}                                   & \textbf{Coded Mean (SD)} & \textbf{Raw Mean (SD)} & \textbf{Moral Relevance}            \\
		\midrule
		Better donor compatibility                         & $0.97 (0.18)^{***}$      & $2.42 (0.95)^{***}$    & Counting for                        \\ 
		Higher chance of organ acceptance                  & $0.98 (0.16)^{***}$      & $2.25 (0.99)^{***}$    & Counting for                        \\ 
		Spent longer on the waiting list                   & $0.95 (0.22)^{***}$      & $2.15 (1.07)^{***}$    & Counting for                        \\ 
		Longer expected lifespan                           & $0.91 (0.29)^{***}$      & $2.06 (1.18)^{***}$    & Counting for                        \\ 
		Expected full recovery                             & $0.90 (0.30)^{***}$      & $1.99 (1.21)^{***}$    & Counting for                        \\ 
		Healthier expected lifestyle                       & $0.88 (0.32)^{***}$      & $1.95 (1.18)^{***}$    & Counting for                        \\ 
		Greater kidney failure                             & $0.98 (0.13)^{***}$      & $1.78 (1.66)^{***}$    & Counting for                        \\ 
		Clearer documented consent                         & $0.88 (0.32)^{***}$      & $1.78 (1.29)^{***}$    & Counting for                        \\ 
		Expected faster recovery                           & $0.91 (0.29)^{***}$      & $1.76 (1.29)^{***}$    & Counting for                        \\ 
		First-time transplant recipient                    & $0.85 (0.36)^{***}$      & $1.76 (1.13)^{***}$    & Counting for                        \\ 
		Better expected adherence to treatment             & $0.87 (0.34)^{***}$      & $1.70 (1.27)^{***}$    & Counting for                        \\ 
		More severe medical condition                      & $0.98 (0.16)^{***}$      & $1.35 (2.08)^{***}$    & Counting for                        \\ 
		Younger                                            & $0.83 (0.37)^{***}$      & $1.30 (1.12)^{***}$    & Counting for                        \\ 
		More urgent health decline                         & $0.98 (0.16)^{***}$      & $1.22 (1.89)^{***}$    & Counting for                        \\ 
		Better health condition                            & $0.87 (0.34)^{***}$      & $0.91 (1.60)^{***}$    & Counting for                        \\ 
		Lower alcohol consumption                          & $0.77 (0.42)^{***}$      & $0.44 (1.45)^{*}$      & Counting for                        \\ 
		Greater responsibility for own illness             & $0.91 (0.29)^{***}$      & $-1.10 (1.69)^{***}$   & Counting against                    \\ 
		History of inhalant abuse & $0.84 (0.37)^{***}$      & $-0.88 (1.71)^{***}$   & Counting against                    \\ 
		Higher tobacco use                                 & $0.82 (0.39)^{***}$      & $-0.88 (1.56)^{***}$   & Counting against                    \\ 
		History of drug misuse                             & $0.82 (0.39)^{***}$      & $-0.87 (1.61)^{***}$   & Counting against                    \\ 
		Higher body weight                                 & $0.67 (0.47)^{**}$       & $-0.43 (1.18)^{**}$    & Counting against                    \\ 
		More underlying medical conditions                 & $0.92 (0.28)^{***}$      & $-0.18 (1.88)^{}$      & Counting either for or against      \\ 
		Stronger ethnic match advantage                    & $0.62 (0.49)^{}$         & $0.92 (1.12)^{***}$    & Divisive                            \\ 
		More caregiving dependents                         & $0.61 (0.49)^{}$         & $0.82 (1.18)^{***}$    & Divisive                            \\ 
		Citizen/legal resident                             & $0.51 (0.50)^{}$         & $0.72 (1.44)^{***}$    & Divisive                            \\ 
		More financial dependents                          & $0.48 (0.50)^{}$         & $0.72 (1.09)^{***}$    & Divisive                            \\ 
		Geographically closer to the transplant center     & $0.53 (0.50)^{}$         & $0.47 (1.27)^{**}$     & Divisive                            \\ 
		Stronger social support system                     & $0.51 (0.50)^{}$         & $0.40 (1.21)^{**}$     & Divisive                            \\ 
		Better financial coverage                          & $0.43 (0.50)^{}$         & $0.38 (1.34)^{*}$      & Divisive                            \\ 
		Religious objections to transplant                 & $0.54 (0.50)^{}$         & $-0.50 (1.48)^{**}$    & Divisive                            \\ 
		Criminal record                                    & $0.42 (0.50)^{}$         & $-0.23 (1.25)^{}$      & \makecell[tl]{Divisive and counting \\either for or against}  \\ 
		Higher social status                             & $0.29 (0.46)^{***}$      & $-0.03 (1.22)^{}$      & Irrelevant                          \\ 
		\bottomrule
	\end{tabular}
\end{table}
\clearpage
\begin{table*}[t]
    \centering
    \small
    \renewcommand{\arraystretch}{0.8}
    \caption{Moderated mediation results (Model 15) for the AI kidney allocation use case (KIDNEY), showing direct and indirect effects of political ideology on moral preferences across three framing conditions. \textit{Note:} An asterisk (*) indicates that the conditional direct effect for conservatives differed significantly from that for progressives. IMM = index of moderated mediation.}
    \label{tab:mm_kidney}
    \begin{tabular}{@{} l l l r r r @{}}
        \toprule
        Moral Feature & Predictor/Mediator & Question Framing & Effect & 95\%(Boot)LCI & 95\%(Boot)UCI \\
        \midrule
        \textbf{Morally Relevant and Positive} & & & & & \\
        \quad More urgent health decline & Conservative $\rightarrow$ Y & Control & 0.33 & -1.17 & 1.83 \\
         &  & \textit{World-You-Want} & 1.86* & 0.20 & 3.53 \\
         &  & Could-Be-You & 0.65 & -0.78 & 2.09 \\
        \addlinespace
        \quad Greater kidney failure & Conservative $\rightarrow$ Y & Control & -0.38 & -1.76 & 1.00 \\
         &  & \textit{World-You-Want} & 1.55* & 0.02 & 3.08 \\
         &  & Could-Be-You & 0.01 & -1.31 & 1.32 \\
        \addlinespace
        \quad First-time transplant recipient & Conservative $\rightarrow$ Y & Control & 0.26 & -0.63 & 1.15 \\
         &  & \textit{World-You-Want} & 1.12* & 0.13 & 2.10 \\
         &  & Could-Be-You & 0.76 & -0.09 & 1.61 \\
        \addlinespace
        \quad Healthier expected lifestyle & Conservative $\rightarrow$ Y & Control & 0.51 & -0.45 & 1.47 \\
         &  & \textit{World-You-Want} & 1.18* & 0.12 & 2.25 \\
         &  & Could-Be-You & 0.53 & -0.39 & 1.44 \\
        \addlinespace
        \quad Expected full recovery & Conservative $\rightarrow$ Y & \textit{Control} & 1.08* & 0.12 & 2.03 \\
         &  & \textit{World-You-Want} & 1.20* & 0.14 & 2.27 \\
         &  & Could-Be-You & -0.03 & -0.94 & 0.89 \\
        \addlinespace
         & Conservative $\rightarrow$ Purity $\rightarrow$ Y & Control & 0.23 & -0.11 & 0.72 \\
         &  & World-You-Want & -0.08 & -0.59 & 0.25 \\
         &  & \quad IMM & -0.31 & -1.07 & 0.12 \\
         &  & \textit{Could-Be-You} & -0.51* & -1.19 & -0.03 \\
         &  & \quad \textit{IMM} & -0.74* & -1.68 & -0.07 \\
        \midrule
        \textbf{Morally Relevant and Negative} & & & & & \\
        \quad History of drug misuse & Conservative $\rightarrow$ Y & Control & 0.28 & -1.00 & 1.57 \\
         &  & World-You-Want & -0.63 & -2.06 & 0.79 \\
         &  & Could-Be-You & 0.26 & -0.97 & 1.49 \\
        \addlinespace
         & Conservative $\rightarrow$ Equality $\rightarrow$ Y & \textit{Control} & -0.67* & -1.44 & -0.11 \\
         &  & World-You-Want & 0.06 & -0.44 & 0.55 \\
         &  & \quad \textit{IMM} & 0.73* & 0.06 & 1.70 \\
         &  & Could-Be-You & -0.13 & -0.71 & 0.33 \\
         &  & \quad IMM & 0.54 & -0.07 & 1.40 \\
        \midrule
        \textbf{Morally Irrelevant} & & & & & \\
        \quad Higher social status & Conservative $\rightarrow$ Y & \textit{Control} & 1.28* & 0.35 & 2.21 \\
         &  & World-You-Want & -0.85 & -1.89 & 0.18 \\
         &  & Could-Be-You & 0.11 & -0.78 & 0.99 \\
        \midrule
        \textbf{Morally Divisive in Relevance} & & & & & \\
        \quad Better financial coverage & Conservative $\rightarrow$ Y & \textit{Control} & 1.29* & 0.22 & 2.36 \\
         &  & World-You-Want & 0.64 & -0.55 & 1.82 \\
         &  & Could-Be-You & 0.83 & -0.19 & 1.85 \\
        \addlinespace
        \quad More underlying medical conditions & Conservative $\rightarrow$ Y & \textit{Control} & -1.71* & -3.20 & -0.23 \\
         &  & World-You-Want & -0.56 & -2.21 & 1.08 \\
         &  & Could-Be-You & 0.61 & -0.81 & 2.02 \\
        \addlinespace
         & Conservative $\rightarrow$ Authority $\rightarrow$ Y & Control & 0.29 & -0.63 & 1.30 \\
         &  & World-You-Want & -0.31 & 1.60 & 0.64 \\
         &  & \quad IMM & -0.61 & -2.25 & 0.73 \\
         &  & \textit{Could-Be-You} & -1.30* & -2.83 & -0.23 \\
         &  & \quad \textit{IMM} & -1.59* & -3.57 & -0.20 \\
        \addlinespace
        \quad & Conservative $\rightarrow$ Purity $\rightarrow$ Y & Control & 0.04 & -0.46 & 0.49 \\
         &  & World-You-Want & 0.54 & -0.11 & 1.40 \\
         &  & \quad IMM & 0.50 & -0.22 & 1.49 \\
         &  & \textit{Could-Be-You} & 0.87* & 0.04 & 2.08 \\
         &  & \quad \textit{IMM} & 0.83* & 0.02 & 2.16 \\
        \bottomrule
    \end{tabular}
\end{table*}
\clearpage
\begin{figure*}[t]
    \centering
    \includegraphics[width=\textwidth]{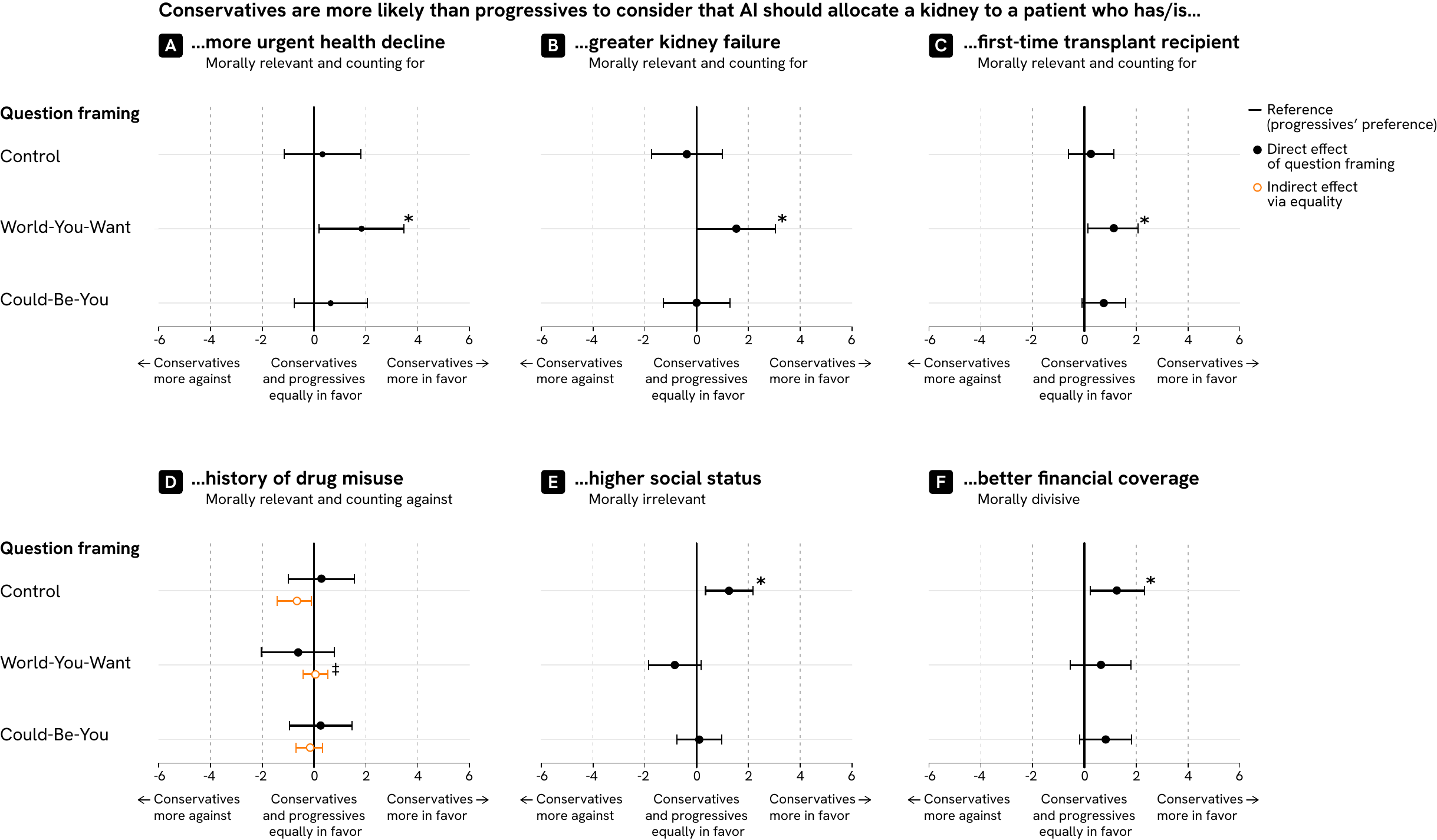}
    \caption{\textbf{Additional moderated mediation results for the AI kidney allocation use case (KIDNEY, Phase 2).} \emph{(A)} In the control and Could-Be-You conditions, no political divide emerged with respect to a patient with more urgent health decline. In the World-You-Want condition, conservatives were more likely than progressives to count for allocating a kidney to this patient. \emph{(B)} In the control and Could-Be-You conditions, conservatives and progressives did not differ in their views about patients with greater kidney failure. However, in the World-You-Want condition, conservatives were more inclined than progressives to support giving the kidney to such a patient. \emph{(C)} In the control and Could-Be-You conditions, no differences appeared between conservatives and progressives regarding first-time transplant recipients. In contrast, under the World-You-Want condition, conservatives were more likely than progressives to favor allocating the kidney to these patients. \emph{(D)} When the patient had a history of drug misuse, no political divide emerged. In the control condition, however, conservatives' weaker endorsement of equality worked against allocating the kidney to this patient. Under the World-You-Want condition, this effect of equality was reduced. \emph{(E)} In the control condition, a political divide emerged for patients with higher social status, with conservatives more supportive of allocating the kidney to them. In both the World-You-Want and Could-Be-You conditions, however, this divide was reduced. \emph{(F)} In the control condition, conservatives were more supportive than progressives of allocating the kidney to patients with better financial coverage. However, this political divide was reduced in both the World-You-Want and Could-Be-You conditions. \textit{Note:} An asterisk (*) indicates that the conditional direct effect for conservatives differed significantly from that for progressives. A double dagger ($\ddagger$) indicates that the index of moderated mediation was statistically significant (bootstrapped 95\% confidence interval excluding zero). See Figure~\ref{fig:study2_model} for the moderated mediation model (Hayes \citeyear{hayes2017introduction}, Model 15).}
    \label{fig:study2:mm:kidney:appen}
\end{figure*}
\clearpage

\section{Phase 1 and Phase 2 Results for Use Case 2: AI Agents Simulating Absent Workers (WORK)}
\label{app:results_work}

\begin{table}[htbp]
    \centering
    \small
    \caption{Moral features identified in Phase 1 as \textit{should be considered} for the AI agents simulating absent workers use case (WORK), color-coded by mention rate: \textcolor{252}{\textbf{medium blue} ($\ge$25\%)}, \textcolor{102}{\textbf{light blue ($\ge$10\%)}}, and \textcolor{02}{\textbf{gray ($<$10\%)}}. Features mentioned by fewer than 2.5\% of participants are omitted.}
    \label{tab:work-study1-should}
    \begin{tabular}{@{\extracolsep{\fill}} p{5.5cm} c c c c}
    \toprule
    \textbf{Moral features} & \textbf{Overall (\%)} & \multicolumn{3}{c}{\textbf{Political ideology(\%)}}\\
    \cmidrule(lr){3-5}
    \textbf{SHOULD BE CONSIDERED}& & \textbf{Progressive} & \textbf{Moderate} & \textbf{Conservative} \\
    \midrule
    \rowcolor{25}Reason for request & 32 & 36 & 26 & 38 \\
    \rowcolor{25}Characteristics of AI agent & 30 & 40 & 35 & 25 \\
    \rowcolor{25}Company policy & 27 & 38 & 28 & 34 \\
    \hline
    \rowcolor{10}Workers personality and intent & 22 & 33 & 33 & 34 \\
    \rowcolor{10}Impact on others and the workplace & 16 & 29 & 46 & 25 \\
    \rowcolor{10}Duration of inactivity & 15 & 26 & 43 & 31 \\
    \rowcolor{10}Fraudulence of activity & 12 & 35 & 35 & 30 \\
    \rowcolor{10}Repercussions for worker & 12 & 17 & 33 & 50 \\
    \rowcolor{10}Consent & 12 & 17 & 39 & 44 \\
    \rowcolor{10}Workers compensation & 11 & 31 & 31 & 38 \\
    \rowcolor{10}Repercussions for third parties & 11 & 27 & 40 & 33 \\
    \hline
    \rowcolor{0}Work context & 10 & 47 & 33 & 20 \\
    \rowcolor{0}Task quality & 9 & 38 & 31 & 31 \\
    \rowcolor{0}Risk to employer & 9 & 31 & 31 & 38 \\
    \rowcolor{0}Type of activity & 9 & 38 & 31 & 31 \\
    \rowcolor{0}Worker's ethical justification & 9 & 31 & 38 & 31 \\
    \rowcolor{0}Employees workload & 8 & 17 & 42 & 41 \\
    \rowcolor{0}Employees productivity & 8 & 42 & 17 & 41 \\
    \rowcolor{0}Worker dealing with emergency & 8 & 42 & 42 & 16 \\
    \rowcolor{0}Workers health status & 7 & 36 & 36 & 28 \\
    \rowcolor{0}Security and privacy of AI & 7 & 45 & 45 & 10 \\
    \rowcolor{0}Frequency of use & 7 & 40 & 30 & 30 \\
    \rowcolor{0}Attendance & 7 & 44 & 44 & 12 \\
    \rowcolor{0}Fairness & 6 & 33 & 56 & 11 \\
    \rowcolor{0}Company context and priorities & 5 & 75 & 0 & 25 \\
    \rowcolor{0}Alternative activity & 5 & 12 & 38 & 50 \\
    \rowcolor{0}Cultural norms & 5 & 57 & 14 & 29 \\
    \rowcolor{0}Personal information & 3 & 20 & 40 & 40 \\
    \rowcolor{0}Task deadline & 3 & 40 & 40 & 20 \\
    \rowcolor{0}Social context & 3 & 25 & 25 & 50 \\
    \rowcolor{0}Benefit to employee & 3 & 0 & 33 & 67 \\
    \bottomrule
\end{tabular}
\end{table}

\newpage

\begin{table}[htbp]
    \centering
    \small
    \caption{Moral features identified in Phase 1 as \textit{should not be considered} for the AI agents simulating absent workers use case (WORK), color-coded by mention rate: \textcolor{102}{\textbf{light blue ($\ge$10\%)}}, and \textcolor{02}{\textbf{gray ($<$10\%)}}. Features mentioned by fewer than 2.5\% of participants are omitted.}
    \label{tab:work-study1-shouldnt}
    \begin{tabular}{@{\extracolsep{\fill}} p{5.5cm} c c c c}
    \toprule
    \textbf{Moral features} & \textbf{Overall (\%)} & \multicolumn{3}{c}{\textbf{Political ideology(\%)}}\\
    \cmidrule(lr){3-5}
    \textbf{SHOULD NOT BE CONSIDERED}& & \textbf{Progressive} & \textbf{Moderate} & \textbf{Conservative} \\
    \midrule
    \rowcolor{10}Characteristics of AI agent & 22 & 21 & 52 & 27 \\
    \rowcolor{10}Workers personality and intent & 21 & 34 & 34 & 32 \\
    \rowcolor{10}Age & 18 & 44 & 19 & 37 \\
    \rowcolor{10}Ethnicity & 18 & 30 & 26 & 44 \\
    \rowcolor{10}Gender & 14 & 41 & 27 & 32 \\
    \rowcolor{10}Personal information & 12 & 63 & 11 & 26 \\
    \rowcolor{10}Social status & 12 & 24 & 41 & 35 \\
    \rowcolor{10}Work context & 11 & 27 & 40 & 33 \\
    \rowcolor{10}Workers compensation & 11 & 27 & 47 & 26 \\
    \hline
    \rowcolor{0}Company policy & 9 & 29 & 36 & 35 \\
    \rowcolor{0}Employee tenure & 9 & 54 & 31 & 15 \\
    \rowcolor{0}Duration of inactivity & 9 & 23 & 46 & 31 \\
    \rowcolor{0}Type of activity & 9 & 54 & 38 & 8 \\
    \rowcolor{0}Employee role & 8 & 42 & 33 & 25 \\
    \rowcolor{0}Reason for request & 7 & 36 & 36 & 28 \\
    \rowcolor{0}Benefit to employee & 7 & 20 & 40 & 40 \\
    \rowcolor{0}Attendance & 6 & 44 & 33 & 23 \\
    \rowcolor{0}Sex & 5 & 38 & 25 & 37 \\
    \rowcolor{0}Company context and priorities & 5 & 29 & 29 & 42 \\
    \rowcolor{0}Sexual orientation & 5 & 57 & 29 & 14 \\
    \rowcolor{0}Time & 5 & 57 & 29 & 14 \\
    \rowcolor{0}Religion & 4 & 17 & 33 & 50 \\
    \rowcolor{0}Task quality & 4 & 67 & 33 & 0 \\
    \rowcolor{0}Work location & 4 & 33 & 50 & 17 \\
    \rowcolor{0}Political ideology & 4 & 0 & 50 & 50 \\
    \rowcolor{0}Employees workload & 3 & 20 & 20 & 60 \\
    \rowcolor{0}Security and privacy of AI & 3 & 0 & 60 & 40 \\
    \rowcolor{0}Workers health status & 3 & 20 & 40 & 40 \\
    \rowcolor{0}Cultural norms & 3 & 0 & 100 & 0 \\
    \rowcolor{0}Employees productivity & 3 & 75 & 0 & 25 \\
    \rowcolor{0}Nationality & 3 & 0 & 50 & 50 \\
    \rowcolor{0}Tiredness & 3 & 0 & 50 & 50 \\
    \rowcolor{0}Past requests & 3 & 25 & 0 & 75 \\
    \rowcolor{0}Impact on others and the workplace & 3 & 50 & 25 & 25 \\
    \bottomrule
\end{tabular}
\end{table}
\clearpage
\begin{table}[t]
    \caption{Moral relevance and importance of features elicited in Phase 2 for the AI agents simulating absent workers use case (WORK). \textit{Note:} $^{***}p<0.001$, $^{**}p<0.01$, $^{*}p<0.05$. Significance levels are based on two separate two-sided one-sample \textit{t}-tests: Mean (Coded) was tested against a reference value of 0.5; Mean was tested against 0.}
    \small
    \label{tab:work-study2}
    \begin{tabular}{llll}
    \toprule
    \textbf{Feature} & \textbf{Coded Mean (SD)} & \textbf{Raw Mean (SD)} & \textbf{Moral Relevance} \\
    \midrule
    Providing a genuine reason & $0.81 (0.40)^{***}$ & $1.48 (1.74)^{***}$ &  Counting for \\
    Emergency situation & $0.82 (0.39)^{***}$ & $1.29 (1.97)^{***}$ &  Counting for \\
    Using AI that aligns with the company's interest & $0.79 (0.41)^{***}$ & $1.16 (1.90)^{***}$ &  Counting for \\
    Doing more work & $0.76 (0.43)^{***}$ & $1.13 (1.70)^{***}$ &  Counting for \\
    Compliant with company policies & $0.75 (0.43)^{***}$ & $1.07 (1.91)^{***}$ &  Counting for \\
    Health problems & $0.64 (0.48)^{*}$ & $1.02 (1.62)^{***}$ &  Counting for \\
    Caregiving responsibilities & $0.68 (0.47)^{**}$ & $1.02 (1.60)^{***}$ &  Counting for \\
    Demonstrating reliability & $0.72 (0.45)^{***}$ & $0.98 (1.67)^{***}$ &  Counting for \\
    Demonstrating productivity & $0.74 (0.44)^{***}$ & $0.95 (1.85)^{***}$ &  Counting for \\
    Using AI that reports actions to the company & $0.78 (0.41)^{***}$ & $0.83 (1.92)^{***}$ &  Counting for \\
    Performing simulatable tasks & $0.76 (0.43)^{***}$ & $0.80 (1.84)^{***}$ &  Counting for \\
    Showing acceptable workplace behavior & $0.70 (0.46)^{***}$ & $0.78 (1.77)^{***}$ &  Counting for \\
    Inactive briefly  ($<$1 hour) & $0.70 (0.46)^{***}$ & $0.60 (1.72)^{**}$ &  Counting for \\
    Facing tight deadlines & $0.68 (0.47)^{**}$ & $0.59 (1.84)^{*}$ &  Counting for \\
    Generating higher profit & $0.69 (0.46)^{***}$ & $0.58 (1.85)^{*}$ &  Counting for \\
    Paid but not doing the work & $0.88 (0.32)^{***}$ & $-0.93 (2.41)^{**}$ &  Counting against \\
    Disrupts workplace operations & $0.88 (0.33)^{***}$ & $-0.84 (2.30)^{**}$ &  Counting against \\
    Sharing sensitive data & $0.85 (0.36)^{***}$ & $-0.74 (2.38)^{*}$ &  Counting against \\
    Misusing tools for fraudulent purposes & $0.89 (0.31)^{***}$ & $-0.72 (2.55)^{*}$ &  Counting against \\
    Putting employer at risk & $0.87 (0.34)^{***}$ & $-0.68 (2.40)^{*}$ &  Counting against \\
    Doing low-impact work & $0.73 (0.44)^{***}$ & $0.28 (1.83)^{}$ &  Counting either for or against \\
    New to company & $0.70 (0.46)^{***}$ & $0.11 (1.86)^{}$ &  Counting either for or against \\
    Using highly autonomous AI & $0.75 (0.43)^{***}$ & $0.02 (2.08)^{}$ &  Counting either for or against \\
    Doing task for first time & $0.64 (0.48)^{*}$ & $0.01 (1.83)^{}$ &  Counting either for or against \\
    Increasing coworkers' workload & $0.82 (0.38)^{***}$ & $-0.57 (2.15)^{}$ &  Counting either for or against \\
    Undermining social/professional norms & $0.86 (0.35)^{***}$ & $-0.53 (2.38)^{}$ &  Counting either for or against \\
    Doing another activity unrelated to work & $0.74 (0.44)^{***}$ & $-0.33 (2.03)^{}$ &  Counting either for or against \\
    Receiving significant benefits & $0.67 (0.47)^{**}$ & $-0.28 (1.90)^{}$ &  Counting either for or against \\
    At risk of firing & $0.73 (0.44)^{***}$ & $-0.27 (2.09)^{}$ &  Counting either for or against \\
    Not seeking manager approval & $0.70 (0.46)^{***}$ & $-0.26 (1.98)^{}$ &  Counting either for or against \\
    Treated unfairly & $0.63 (0.48)^{}$ & $0.52 (1.79)^{*}$ & Divisive \\
    Having high morale & $0.52 (0.50)^{}$ & $0.45 (1.51)^{*}$ & Divisive \\
    Able to request time off & $0.63 (0.48)^{}$ & $0.28 (1.83)^{}$ & \makecell[tl]{Divisive and counting\\either for or against} \\
    In a healthy work environment & $0.59 (0.49)^{}$ & $0.12 (1.72)^{}$ & \makecell[tl]{Divisive and counting\\either for or against} \\
    External contractor & $0.60 (0.49)^{}$ & $-0.16 (1.73)^{}$ & \makecell[tl]{Divisive and counting\\either for or against} \\
    \bottomrule
  \end{tabular}
\end{table}
\clearpage
\begin{table*}[t]
  \centering
  \caption{Moderated mediation results (Model 15) for the AI agents simulating absent workers use case (WORK), showing direct and indirect effects of political ideology on moral preferences across three framing conditions. \textit{Note:} An asterisk (*) indicates that the conditional direct effect for conservatives differed significantly from that for progressives. IMM = index of moderated mediation.}
  
  \small
  \label{tab:mm_work}
  \resizebox{\linewidth}{!}{%
  \begin{tabular}{lllrrr}
    \toprule
    Moral Feature & Predictor/Mediator & Question Framing & Effect & 95\%(Boot)LCI & 95\%(Boot)UCI \\
    \midrule
    \textbf{Morally Relevant and Positive} & & & & & \\
    \quad Experiencing health problems & Conservative $\rightarrow$ Y & Control & 0.31 & -1.42 & 2.04 \\
     &  & World-You-Want & 1.19 & -0.38 & 2.77 \\
     &  & Could-Be-You & 1.01 & -0.49 & 2.51 \\
     \addlinespace
     & Conservative $\rightarrow$ Loyalty $\rightarrow$ Y & Control & -0.20 & -0.78 & 0.19 \\
     &  & \textit{World-You-Want} & 0.98* & 0.13 & 2.00 \\
      &  & \multicolumn{1}{r}{\textit{IMM}} & 1.18* & 0.24 & 2.35 \\
     &  & Could-Be-You & -0.28 & -1.02 & 0.38 \\
     &  & \multicolumn{1}{r}{IMM} & -0.08 & -0.64 & 0.39 \\
     \addlinespace
     & Conservative $\rightarrow$ Authority $\rightarrow$ Y & Control & 0.42 & -0.90 & 1.43 \\
     &  & \textit{World-You-Want} & -2.24* & -4.14 & -0.79 \\
          &  & \multicolumn{1}{r}{\textit{IMM}} & -2.66* & -4.85 & -0.92 \\
     &  & Could-Be-You & 0.40 & -1.03 & 1.67 \\
     &  & \multicolumn{1}{r}{IMM} & -0.01 & -0.79 & 0.65 \\
     \midrule
    \textbf{Morally Relevant and Negative} & & & & & \\
    \quad Being paid but not working & Conservative $\rightarrow$ Y & \textit{Control} & -2.79* & -5.58 & -0.00 \\
     &  & World-You-Want & -0.43 & -2.96 & 2.10 \\
     &  & Could-Be-You & 0.01 & -2.41 & 2.43 \\
     \midrule
    \textbf{Morally Divisive in Direction} & & & & & \\
    \quad Receiving significant benefits & Conservative $\rightarrow$ Y & Control & -0.04 & -2.14 & 2.06 \\
     &  & World-You-Want & 0.45 & -1.45 & 2.36 \\
     &  & Could-Be-You & 1.04 & -0.78& 2.86 \\
     \addlinespace
     & Conservative $\rightarrow$ Loyalty $\rightarrow$ Y & Control & -0.17 & -0.78 & 0.54 \\
     &  & \textit{World-You-Want} & 0.99* & 0.18 & 2.85 \\
          &  & \multicolumn{1}{r}{\textit{IMM}} & 1.16* & 0.21 & 2.62 \\
     &  & Could-Be-You & 0.37 & -0.46 & 1.20 \\
     &  & \multicolumn{1}{r}{IMM} & 0.53 & -0.19 & 1.35 \\
    \bottomrule
  \end{tabular}
  }
\end{table*}
\clearpage

\section{Phase 1 and Phase 2 Results for Use Case 3: Generative AI Content of the Deceased (GEN)}
\label{app:results_art}

\begin{table}[htbp]
    \centering
    \small
    \caption{Moral features identified in Phase 1 as \textit{should be considered} for the generative AI content of the deceased use case (GEN), color-coded by mention rate: \textcolor{252}{\textbf{medium blue} ($\ge$25\%)}, \textcolor{102}{\textbf{light blue ($\ge$10\%)}}, and \textcolor{02}{\textbf{gray ($<$10\%)}}. Features mentioned by fewer than 2.5\% of participants are omitted.}
    \label{tab:art-study1-should}
    \begin{tabular}{@{\extracolsep{\fill}} p{5.5cm} c c c c}
    \toprule
    \textbf{Moral features} & \textbf{Overall (\%)} & \multicolumn{3}{c}{\textbf{Political ideology(\%)}}\\
    \cmidrule(lr){3-5}
    \textbf{\textit{SHOULD BE CONSIDERED}}& & \textbf{Progressive} & \textbf{Moderate} & \textbf{Conservative} \\
    \midrule
    \rowcolor{25}Intended purpose of the video & 40 & 36 & 31 & 33 \\
    \rowcolor{25}Deceased consent & 28 & 34 & 34 & 32 \\
    \rowcolor{25}Video quality & 25 & 45 & 24 & 31 \\
    \hline
    \rowcolor{10}Relation to deceased & 24 & 54 & 19 & 27 \\
    \rowcolor{10}Risk of harm by/misuse of video & 23 & 31 & 37 & 32 \\
    \rowcolor{10}Distress to friends and family of deceased & 19 & 17 & 48 & 35 \\
    \rowcolor{10}Family consent & 19 & 31 & 41 & 28 \\
    \rowcolor{10}Age of deceased & 16 & 23 & 32 & 45 \\
    \rowcolor{10}Third-party consent & 15 & 43 & 30 & 27 \\
    \rowcolor{10}Death circumstances & 14 & 19 & 43 & 38 \\
    \rowcolor{10}Motivation behind the video & 14 & 43 & 29 & 28 \\
    \rowcolor{10}Legality of request & 12 & 42 & 32 & 26 \\
    \rowcolor{10}Realism of the video & 12 & 33 & 44 & 23 \\
    \rowcolor{10}Cultural acceptability of request & 10 & 31 & 38 & 31 \\
    \rowcolor{10}Distress to viewers & 10 & 31 & 44 & 25 \\
    \hline
    \rowcolor{0}Financial motivation of request & 8 & 31 & 38 & 31 \\
    \rowcolor{0}Intended audience & 8 & 20 & 30 & 50 \\
    \rowcolor{0}Appearance of deceased & 6 & 11 & 33 & 56 \\
    \rowcolor{0}Identity of requester & 6 & 44 & 33 & 23 \\
    \rowcolor{0}Deceased public and personal record & 6 & 33 & 33 & 34 \\
    \rowcolor{0}Technical feasibility & 5 & 38 & 12 & 50 \\
    \rowcolor{0}Accuracy of representation & 5 & 14 & 57 & 29 \\
    \rowcolor{0}Age of requester & 5 & 12 & 12 & 76 \\
    \rowcolor{0}Deceased's social media popularity& 5 & 38 & 38 & 24 \\
    \rowcolor{0}Fame of deceased & 5 & 38 & 25 & 37 \\
    \rowcolor{0}Privacy of deceased & 5 & 57 & 43 & 0 \\
    \rowcolor{0}Religion of the deceased & 4 & 60 & 40 & 0 \\
    \rowcolor{0}Mental health & 4 & 50 & 33 & 17 \\
    \rowcolor{0}Deceased identity & 3 & 25 & 0 & 75 \\
    \rowcolor{0}Social context & 3 & 33 & 33 & 34 \\
    \rowcolor{0}Criminal record of deceased & 3 & 0 & 25 & 75 \\
    \rowcolor{0}Commercial use & 3 & 67 & 33 & 0 \\
    \rowcolor{0}Data privacy & 3 & 75 & 25 & 0 \\
\bottomrule
\end{tabular}
\end{table}

\newpage

\begin{table}[htbp]
    \centering
    \small
    \caption{Moral features identified in Phase 1 as \textit{should not be considered} for the generative AI content of the deceased use case (GEN), color-coded by mention rate: \textcolor{102}{\textbf{light blue ($\ge$10\%)}} and \textcolor{02}{\textbf{gray ($<$10\%)}}. Features mentioned by fewer than 2.5\% of participants are omitted.}
    \label{tab:art-study1-shouldnt}
    \begin{tabular}{@{\extracolsep{\fill}} p{5.5cm} c c c c}
    \toprule
    \textbf{Moral features} & \textbf{Overall (\%)} & \multicolumn{3}{c}{\textbf{Political ideology(\%)}}\\
    \cmidrule(lr){3-5}
    \textbf{\textit{SHOULD NOT BE CONSIDERED}}& & \textbf{Progressive} & \textbf{Moderate} & \textbf{Conservative} \\
    \midrule
    \rowcolor{10}Financial motivation of request & 18 & 36 & 43 & 21 \\
    \rowcolor{10}Fame of deceased & 16 & 42 & 38 & 20 \\
    \rowcolor{10}Video quality & 14 & 48 & 43 & 9 \\
    \rowcolor{10}Age of deceased & 13 & 30 & 50 & 20 \\
    \rowcolor{10}Social status of deceased & 12 & 21 & 32 & 47 \\
    \rowcolor{10}Identity of requester & 12 & 50 & 17 & 33 \\
    \rowcolor{10}Technical feasibility & 12 & 22 & 39 & 39 \\
    \rowcolor{10}Gender of deceased & 12 & 28 & 28 & 44 \\
    \rowcolor{0}Ethnicity of deceased & 12 & 33 & 33 & 34 \\
    \rowcolor{10}Deceased's social media popularity & 10 & 40 & 33 & 27 \\
    \rowcolor{10}Appearance of deceased & 10 & 23 & 46 & 31 \\
    \hline
    \rowcolor{0}Intended purpose of the video & 8 & 58 & 25 & 17 \\    
    \rowcolor{0}Length of the video & 8 & 42 & 33 & 25 \\
    \rowcolor{0}Geographic location & 6 & 30 & 60 & 10 \\
    \rowcolor{0}Deceased public and personal record & 6 & 33 & 0 & 67 \\
    \rowcolor{0}Realism of the video & 6 & 22 & 56 & 22 \\
    \rowcolor{0}Social context & 5 & 25 & 62 & 13 \\
    \rowcolor{0}Requester identity & 5 & 57 & 14 & 29 \\
    \rowcolor{0}Deceased consent & 5 & 29 & 57 & 14 \\
    \rowcolor{0}Death circumstances & 5 & 33 & 33 & 34 \\
    \rowcolor{0}Religion of the deceased & 5 & 29 & 43 & 28 \\
    \rowcolor{0}Legality of request & 4 & 33 & 0 & 67 \\
    \rowcolor{0}Sexual orientation & 4 & 17 & 0 & 83 \\
    \rowcolor{0}Political ideology & 4 & 33 & 50 & 17 \\
    \rowcolor{0}Data privacy & 4 & 50 & 33 & 17 \\
    \rowcolor{0}Cost of video & 3 & 0 & 20 & 80 \\
    \rowcolor{0}Time of request & 3 & 60 & 0 & 40 \\
    \rowcolor{0}Third-party consent & 3 & 20 & 40 & 40 \\
    \rowcolor{0}Privacy of deceased & 3 & 0 & 0 & 100 \\
    \rowcolor{0}Ease of video generation & 3 & 75 & 0 & 25 \\
    \rowcolor{0}Motivation behind the video & 3 & 25 & 25 & 50 \\
    \rowcolor{0}Emotional state of requester & 3 & 50 & 25 & 25 \\
\bottomrule
\end{tabular}
\end{table}
\clearpage
\begin{table}[t]
  \caption{Moral relevance and importance of features elicited in Phase 2 for the generative AI content of the deceased use case (GEN). \textit{Note:} $^{***}p<0.001$, $^{**}p<0.01$, $^{*}p<0.05$. Significance levels are based on two separate two-sided one-sample \textit{t}-tests: Mean (Coded) was tested against a reference value of 0.5; Mean was tested against 0.}
  
  \label{tab:art-study2}
  \small
  \setlength{\tabcolsep}{4pt}
  \begin{tabularx}{\textwidth}{Xlll}
    \toprule
    \textbf{Feature} & \textbf{Coded Mean (SD)} & \textbf{Raw Mean (SD)} & \textbf{Moral Relevance} \\
    \midrule
    The video is legally compliant & $0.96 (0.20)^{***}$ & $2.25 (1.42)^{***}$ &  Counting for \\
    The artist has family's consent & $0.95 (0.22)^{***}$ & $2.19 (1.40)^{***}$ &  Counting for \\
    The video honors the deceased & $0.92 (0.26)^{***}$ & $2.18 (1.33)^{***}$ &  Counting for \\
    The artist has deceased's explicit consent & $0.96 (0.20)^{***}$ & $2.08 (1.65)^{***}$ &  Counting for \\
    The artist has other relevant party's consent & $0.92 (0.28)^{***}$ & $2.07 (1.26)^{***}$ &  Counting for \\
    The artist is respectful to the deceased & $0.90 (0.30)^{***}$ & $1.91 (1.57)^{***}$ &  Counting for \\
    The video is for grief support/memorial purpose & $0.89 (0.31)^{***}$ & $1.88 (1.49)^{***}$ &  Counting for \\
    The video is respectful of the deceased's tradition & $0.92 (0.28)^{***}$ & $1.87 (1.58)^{***}$ &  Counting for \\
    The deceased is depicted non-sexually & $0.92 (0.28)^{***}$ & $1.80 (1.75)^{***}$ &  Counting for \\
    The video aligns with deceased's religion & $0.89 (0.31)^{***}$ & $1.78 (1.56)^{***}$ &  Counting for \\
    The artist is a close friend/relative of the deceased & $0.86 (0.35)^{***}$ & $1.78 (1.35)^{***}$ &  Counting for \\
    The video is for non-commercial purpose & $0.87 (0.34)^{***}$ & $1.78 (1.27)^{***}$ &  Counting for \\
    \makecell[tl]{The artist has more transparent or\\understandable motivation} & $0.82 (0.38)^{***}$ & $1.72 (1.29)^{***}$ &  Counting for \\
    The video accurately represents the deceased & $0.89 (0.31)^{***}$ & $1.70 (1.60)^{***}$ &  Counting for \\
    The video keeps sensitive information private & $0.93 (0.25)^{***}$ & $1.66 (1.91)^{***}$ &  Counting for \\
    The video is respectful of community norms & $0.79 (0.41)^{***}$ & $1.53 (1.51)^{***}$ &  Counting for \\
    The deceased has uncontroversial cause of death & $0.81 (0.40)^{***}$ & $1.18 (1.67)^{***}$ &  Counting for \\
    The deceased is public or historical figure & $0.79 (0.41)^{***}$ & $1.18 (1.57)^{***}$ &  Counting for \\
    The video is intended for adults & $0.75 (0.43)^{***}$ & $0.98 (1.81)^{***}$ &  Counting for \\
    The video has wider emotional impact & $0.74 (0.44)^{***}$ & $0.68 (1.74)^{***}$ &  Counting for \\
    The video is intended for public sharing & $0.90 (0.30)^{***}$ & $0.67 (2.12)^{**}$ &  Counting for \\
    The client has high mental health risk & $0.85 (0.36)^{***}$ & $0.38 (2.27)^{}$ &  Counting either for or against \\
    The video may cause potential distress to friend/family & $0.98 (0.16)^{***}$ & $-0.47 (2.70)^{}$ &  Counting either for or against \\
    \makecell[tl]{The video has risk of misleading or\\misrepresenting the deceased} & $0.98 (0.16)^{***}$ & $-0.45 (2.75)^{}$ &  Counting either for or against \\
    The artist is underage & $0.92 (0.26)^{***}$ & $-0.28 (2.52)^{}$ &  Counting either for or against \\
    The video is easy to create & $0.60 (0.49)^{}$ & $0.85 (1.47)^{***}$ & Divisive \\
    The artist is working for oneself & $0.53 (0.50)^{}$ & $0.74 (1.36)^{***}$ & Divisive \\
    The video is about an unmarried deceased & $0.40 (0.49)^{}$ & $0.66 (1.20)^{***}$ & Divisive \\
    The deceased died recently (less than 100 yrs) & $0.62 (0.49)^{}$ & $0.65 (1.55)^{***}$ & Divisive \\
    The deceased has a criminal record & $0.58 (0.50)^{}$ & $0.31 (1.61)^{}$ & \makecell[tl]{Divisive and counting\\either for or against} \\
    \bottomrule
  \end{tabularx}
\end{table}
\clearpage
\begin{table*}[t]
  \centering
  \caption{Moderated mediation results (Model 15) for the generative AI content of the deceased use case (GEN), showing direct and indirect effects of political ideology on moral preferences across three framing conditions. \textit{Note:} An asterisk (*) indicates that the conditional direct effect for conservatives differed significantly from that for progressives. IMM = index of moderated mediation.}
  \label{tab:mm_art}
  \small
  \renewcommand{\arraystretch}{0.9}
  \resizebox{\linewidth}{!}{
  \begin{tabular}{lllrrr}
    \toprule
    Moral Feature & Predictor/Mediator & Question Framing & Effect & 95\%(Boot)LCI & 95\%(Boot)UCI \\
    \midrule
    \textbf{Morally Relevant and Positive} & & & & & \\
    \quad The video is for grief support  & Conservative $\rightarrow$ Y & Control & 0.33 & -1.16 & 1.82 \\
    \quad or memorial purpose &  & World-You-Want & -0.33 & -1.59 & 0.93 \\
     &  & \textit{Could-Be-You} & 2.32** & 0.94 & 3.71 \\
     \addlinespace
    \quad The artist is close friend/relative & Conservative $\rightarrow$ Y & Control & -0.14 & -1.55 & 1.27 \\
     &  & World-You-Want & -0.85 & -2.04 & 0.34 \\
     &  & \textit{Could-Be-You} & 1.56* & 0.25 & 2.86 \\
     \addlinespace
    \quad The video is intended for adults & Conservative $\rightarrow$ Y & \textit{Control} & -1.99* & -3.85 & -0.14 \\
     &  & World-You-Want & 0.49 & -1.08 & 2.05 \\
     &  & Could-Be-You & 0.19 & -1.53 & 1.91 \\
     \addlinespace
    \quad The video is respectful  & Conservative $\rightarrow$ Y & Control & -1.50 & -3.03 & 0.03 \\
     \quad of deceased's tradition &  & World-You-Want & -0.55 & -1.85 & 0.74 \\
     &  & Could-Be-You & 0.90 & -0.51 & 2.32 \\
     \addlinespace
     & Conservative $\rightarrow$ Authority $\rightarrow$ Y & \textit{Control} & 1.49* & 0.28 & 2.98 \\
     &  & World-You-Want & -0.58 & -1.55 & 0.38 \\
       &  & \multicolumn{1}{r}{\textit{IMM}} & -2.08* & -3.95 & -0.48 \\
     &  & Could-Be-You & -0.01 & -1.31 & 1.22 \\
     &  & \multicolumn{1}{r}{IMM} & -1.51 & -3.48 & 0.16 \\
     \addlinespace
    \quad The video aligns with  & Conservative $\rightarrow$ Y & Control & 0.18 & -1.38 & 1.74 \\
     \quad deceased's religion &  & World-You-Want & 0.09 & -1.22 & 1.41 \\
     &  & Could-Be-You & -0.40 & -1.84 & 1.05 \\
     \addlinespace
     & Conservative $\rightarrow$ Authority $\rightarrow$ Y & \textit{Control} & 2.03** & 0.82 & 3.75 \\
     &  & World-You-Want & -0.14 & -0.98 & 0.85 \\
        &  & \multicolumn{1}{r}{\textit{IMM}} & -2.17* & -4.12 & -0.56 \\
     &  & Could-Be-You & 0.41 & -0.72 & 1.82 \\
     &  & \multicolumn{1}{r}{IMM} & -1.63 & -3.66 & 0.15 \\
     \addlinespace
     & Conservative $\rightarrow$ Purity $\rightarrow$ Y & \textit{Control} & -1.29* & -2.65 & -0.34 \\
     &  & World-You-Want & 0.27 & -0.40 & 1.09 \\
          &  & \multicolumn{1}{r}{\textit{IMM}} & 1.56* & 0.44 & 3.14 \\
     &  & Could-Be-You & -0.22 & -1.08 & 0.55 \\
     &  & \multicolumn{1}{r}{IMM} & 1.06 & -0.10 & 2.60 \\
     \addlinespace
    \quad The video honors the deceased & Conservative $\rightarrow$ Y & Control & 0.59 & -0.58 & 1.76 \\
     &  & World-You-Want & 0.07 & -0.92 & 1.06 \\
     &  & Could-Be-You & 0.17 & -0.92 & 1.25 \\
     \addlinespace
     & Conservative $\rightarrow$ Authority $\rightarrow$ Y & \textit{Control} & 0.94* & 0.18 & 2.20 \\
     &  & World-You-Want & -0.31 & -1.02 & 0.38 \\
      &  & \multicolumn{1}{r}{\textit{IMM}} & -1.25* & -2.69 & -0.17 \\
     &  & Could-Be-You & 0.26 & -0.76 & 1.41 \\
     &  & \multicolumn{1}{r}{IMM} & -0.68 & -2.29 & 0.68 \\
     \addlinespace
     & Conservative $\rightarrow$ Purity $\rightarrow$ Y & \textit{Control} & -1.32** & -2.42 & -0.54 \\
     &  & World-You-Want & -0.06 & -0.57 & 0.54 \\
         &  & \multicolumn{1}{r}{\textit{IMM}} & 1.26* & 0.40 & 2.50 \\
     &  & Could-Be-You & 0.05 & -0.88 & 0.80 \\
     &  & \multicolumn{1}{r}{\textit{IMM}} & 1.36* & 0.26 & 2.75 \\
     \midrule
    \textbf{Morally Divisive in Relevance} & & & & & \\
    \quad The video has risk of misleading & Conservative $\rightarrow$ Y & \textit{Control} & -2.91* & -5.62 & -0.20 \\
    \quad or misrepresenting the deceased&  & World-You-Want & 1.06 & -1.23 & 3.35 \\
     &  & Could-Be-You & -1.88 & -4.40 & 0.63 \\
     \addlinespace
    \quad The video may cause potential  & Conservative $\rightarrow$ Y & \textit{Control} & -2.63* & -5.25 & -0.01 \\
     \quad  distress to friend or family &  & World-You-Want & -0.16 & -2.37 & 2.06 \\
     &  & Could-Be-You & -1.35 & -3.78 & 1.08 \\
    \bottomrule
  \end{tabular}%
  }
\end{table*}
\clearpage
\begin{figure*}[t]
    \centering
    \includegraphics[width=\textwidth]{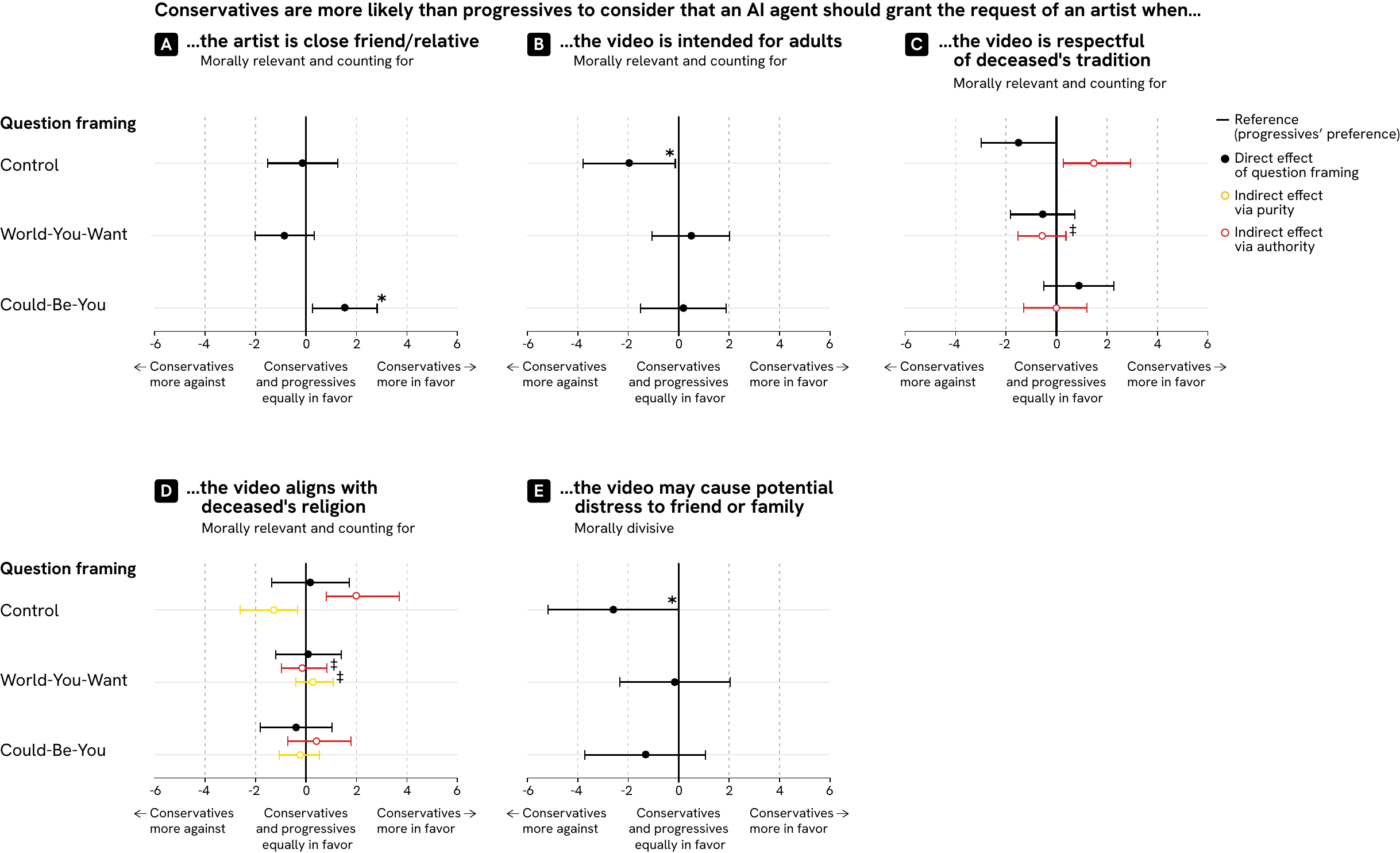}
    \caption{\textbf{Additional moderated mediation results for the generative AI content of the deceased use case (GEN, Phase 2).} \emph{(A)} When the artist was a close friend or relative of the deceased, no political divide emerged in either the control or World-You-Want conditions. In the Could-Be-You condition, however, conservatives were more likely than progressives to support granting the artist's request to create the video. \emph{(B)} In the control condition, a political divide emerged when the video was intended for adults, with conservatives less likely than progressives to grant the artist's request. This divide was mitigated under both question framings. \emph{(C)} When the video was respectful of the deceased's tradition, no political divide emerged. In the control condition, the conditional indirect effect via authority contributed to granting the request, but this effect was diminished under the World-You-Want framing. \emph{(D)} When the video aligned with the deceased's religion, no political divide emerged. In the control condition, however, a moral tension between authority and purity was observed: the conditional indirect effect via authority contributed to granting the request, whereas the conditional indirect effect via purity contributed against granting it. This tension was diminished under the World-You-Want framing. \emph{(E)} In the control condition, a political divide emerged when the video was likely to cause distress to the deceased's friends or family. Conservatives were less likely than progressives to grant the artist's request in this context. This divide was mitigated under both question framings. \textit{Note:} An asterisk (*) indicates that the conditional direct effect for conservatives differed significantly from that for progressives. A double dagger ($\ddagger$) indicates that the index of moderated mediation was statistically significant (bootstrapped 95\% confidence interval excluding zero). See Figure~\ref{fig:study2_model} for the moderated mediation model (Hayes \citeyear{hayes2017introduction}, Model 15).}
    \label{fig:study2:mm:art:appen}
\end{figure*}
\clearpage

\end{document}